\documentclass[twoside]{article}

\usepackage[accepted]{aistats2023}

\usepackage[utf8]{inputenc} 
\usepackage[T1]{fontenc}    
\usepackage[hyphens]{url}
\usepackage{hyperref}
\usepackage[hyphenbreaks]{breakurl}
\usepackage{booktabs}       
\usepackage{amsfonts}       
\usepackage{nicefrac}       
\usepackage{microtype}      
\usepackage{xcolor}         

\usepackage{graphicx}
\usepackage{amsmath}
\usepackage{amssymb}
\usepackage{amsthm}
\usepackage{bbm}
\usepackage{xspace}
\usepackage{footnote}
\usepackage{amsfonts}

\usepackage{hhline}
\usepackage{multirow}
\usepackage{setspace} 
\usepackage{epsfig}

\usepackage{booktabs}
\usepackage{xcolor}
\usepackage{lipsum}
\usepackage{listings}
\usepackage{colortbl}
\usepackage{arydshln} 
\usepackage{makecell} 
\usepackage{tikz}
\usepackage{varwidth}
\usepackage{natbib}
\usepackage{algorithm}
\usepackage{algorithmic}
\usepackage{subfig}

\usetikzlibrary{fit,arrows,calc,positioning,quotes}
\usetikzlibrary{decorations.pathreplacing}

\usepackage{mathtools}
\usepackage{fancyvrb}
\usepackage{textcomp}
\usepackage{stmaryrd}
\usepackage{pgfplots}
\usepackage{pgfplotstable}

\newtheorem{theorem}{Theorem}[section]
\newtheorem{lemma}[theorem]{Lemma}
\newtheorem{definition}[theorem]{Definition}

\tikzset{
  -|-/.style={
    to path={
      (\tikztostart) -| ($(\tikztostart)!#1!(\tikztotarget)$) |- (\tikztotarget)
      \tikztonodes
    }
  },
  -|-/.default=0.5,
  |-|/.style={
    to path={
      (\tikztostart) |- ($(\tikztostart)!#1!(\tikztotarget)$) -| (\tikztotarget)
      \tikztonodes
    }
  },
  |-|/.default=0.5
}

\usetikzlibrary{shapes.geometric}

\long\def\com#1{}

\long\def\ruzica#1{{\color{red}{\bf Ruzica: }{\small [#1]}}}

\newcommand{\pathimprovement}{\ensuremath{63.38\%}\xspace}
\newcommand{\casdepth}{\ensuremath{1.93}\xspace}
\newcommand{\classicdepth}{\ensuremath{5.27}\xspace}

\newcommand{\trainingSet}{\ensuremath{\mathcal{S}}\xspace}

\newcommand{\para}[1]{\smallskip\noindent {\bf #1}}

\newcommand{\squishlist}{
   \begin{list}{$\bullet$}
    { \setlength{\itemsep}{0pt}      \setlength{\parsep}{3pt}
      \setlength{\topsep}{3pt}       \setlength{\partopsep}{0pt}
      \setlength{\leftmargin}{3.5mm} \setlength{\labelwidth}{1em}
      \setlength{\labelsep}{0.5em} } 
}

\newcommand{\squishend}{
    \end{list}  }

\def\BibTeX{{\rm B\kern-.05em{\sc i\kern-.025em b}\kern-.08em
    T\kern-.1667em\lower.7ex\hbox{E}\kern-.125emX}}

\begin{document}

\twocolumn[

\aistatstitle{Succinct Explanations with Cascading Decision Trees}
\aistatsauthor{ Jialu Zhang \And Yitan Wang \And  Mark Santolucito \And Ruzica Piskac }

\aistatsaddress{ Yale University \And  Yale University \And Barnard College \And Yale University } 
]

\begin{abstract}

The decision tree is one of the most popular and classical machine learning models from the 1980s. 
However, in many practical applications, decision trees tend to generate decision paths with excessive depth. 
Long decision paths often cause overfitting problems, and make models difficult to interpret. 
With longer decision paths, inference is also more likely to fail when the data contain missing values. 
In this work, we propose a new tree model called Cascading Decision Trees to alleviate this problem. 
The key insight of Cascading Decision Trees is to separate the decision path and the explanation path. 
Our experiments show that on average, Cascading Decision Trees generate \pathimprovement shorter explanation paths, avoiding overfitting and thus achieve higher test accuracy. 
We also empirically demonstrate that Cascading Decision Trees have advantages in the robustness against missing values.
\end{abstract}


\section{Introduction}
\label{sec:intro}

Binary classification is the process of classifying the given input set into two classes based on some
classification criteria. Binary classification is widely used in everyday life: for 
example, a typical application for binary classification is determining whether a patient has some 
disease by analyzing their comprehensive medical record.
Existing work on binary classification mainly uses the accuracy of prediction as the main criterion for evaluating model performance.
However, in order for a model to be useful in real-world applications, it is imperative that users are able to understand and explain the logic underlying model predictions.
In many real-world applications, especially the medical and scientific domains, model comprehensibility\footnote{In this paper, comprehensibility and interpretability are used interchangeably.} is of the utmost importance.
In these cases, users need to understand the classification model to scientifically explain the reasons behind the classification or even rely on the model itself to discover a possible solution to the target problem.

There is a trade-off between enough explainability and high classification accuracy using current models.
``Black-box'' models such as deep neural networks, random forests, and ensemble methods tend to have the highest accuracy in binary classification~\cite{comprehensible14freitas,explainable18}. 
However, their opaque structure hinders understandability, making the logic behind the predictions difficult to trace.
This lack of transparency may further discourage the use of the model~\cite{reverse12augasta,seeing07van}. 

Decision tree models, on the other hand, have transparent decision making steps.
A traversal of features on the decision path from the root to the leaf node is presented to users as a rule.
Therefore, compared to other models, the decision trees models have historically been characterized as having high comprehensibility~\cite{comprehensible14freitas,explainable18}.
However, whether models generated by classic decision trees provide enough comprehensibility 
has been challenged. 
Decision trees can grow large enough that no human could verify their correctness~\cite{caruana1999case}.
Otherwise, they may contain subtrees with redundant attribute conditions, resulting in misinterpretation of the root cause for model predictions~\cite{comprehensible14freitas}.

A succinct explanation also helps to avoid the overfitting problem.
For the decision tree, the VC dimension grows exponentially with the depth of the tree~\cite{shalev14understanding}.
By the Fundamental Theorem of Learning Theory\cite{shalev14understanding}, the sample complexity for achieving probably approximately correct (PAC) learnability is proportional to the VC dimension.
Combining these two facts, the sample complexity is exponential with respect to the depth of the decision tree.
If the depth is not constrained, the exponential sample complexity usually leads to the overfitting problem in practice, because often it is the case that the size of training dataset cannot match 
such sample complexity.
The depth is also highly related to the explanation length.
Therefore, unless the volume of training data is large enough, which is usually not true, a succinct explanation for the model is favored.

This work introduces a new training procedure for deriving succinct explanations for positive classifications, while maintaining the overall prediction accuracy. 
We target explanations for positive classifications\footnote{Notice that succinct explanations for negative classifications can be achieved symmetrically by our algorithm.}, based on real-world motivation.
In the medical domain, a positive classification result implies the presence of the disease~\cite{NIHdic}.
The explanation for positive classification indicates the cause of the disease or some additional lab tests needed for a full 
diagnosis~\cite{CDCpositive}.
To this end, we introduce a novel {\emph {Cascading Decision Trees}} model. A Cascading Decision Trees model is a sequence of 
several smaller decision trees (subtrees) with a predefined tree depth. 
The sequence ends when the subtree does not contain any leaves describing positively classified samples. 
Figure~\ref{fig:casdtree} depicts one example of such a Cascading Decision Trees model. 

The idea behind Cascading Decision Trees is that, while most algorithms for building decision trees are greedy and try to classify as many samples as soon as possible, such classifications result in large explanation paths
for the samples in the lower levels of the tree. 
Instead, we construct a subtree of a predefined depth.
That subtree contains a short explanation for the samples that it managed to classify. 
However, a subtree with a predefined depth may misclassify some samples. 
The next step of the Cascading Decision Trees algorithm is to repeat the process on the training set with the samples that were previously classified as positive removed from the set.
This way, the samples classified as positive in the second subtree 
will have a much shorter explanation path than they would in the original decision tree.
In Cascading Decision Trees, an explanation path for a positively classified sample is the path that starts at the root of the corresponding subtree.

\com{
We target explanations for only positive classifications, based on real-world motivation.
In the medical domain, a positive classification result indicates that a person has the disease for which the test is being done~\cite{NIHdic}.
The positive classification is also combined with additional testings needed for a full 
diagnosis~\cite{CDCpositive}. 
Note that, if a practical application arises, our cascading decision trees model could easily be changed to target the
negative classifications.
}

Reducing the size and the depth of a decision tree to improve comprehensibility
has been studied, both from a theoretical
and a practical perspective. However, constructing such optimally small decision trees is an NP-complete problem~\cite{constructing76Laurent}, and the main drawback of 
these approaches is that the model is computationally too expensive to train.
Even when using state-of-the-art libraries~\cite{AglinNS20,VerwerZ19}, we observed the long running times due to high complexity.
To illustrate this on an example, to learn a model on the Ionosphere dataset (from the UCI Machine Learning Repository), the BinOCT tool needs approximately 10 minutes, while our approach completes this task in 1.1 seconds.

We demonstrate the applicability of the Cascading Decision Trees model in two ways.
First, we use our model to perform binary classification on ten benchmarks from UCI Machine Learning Repository, which are standard benchmarks for comparing the performance of tree-based classification methods~\cite{constrained04zhao}.
Second, we apply our model to a new application, namely continuous integration (CI) build status prediction~\cite{santolucito2018statically}.
Overall, we report that compared to decision tree with PI-explanation~\cite{ShihCD18,izza2020explaining}, 
our approach shortens the explanation depth for positive classifications by an average of \pathimprovement while still maintaining the prediction accuracy.

\section{Related Work}


\para{Explainable Artificial Intelligence: XAI.}
The comprehensibility of classification models~\cite{ShihCD18,explainable18,NEURIPS2019_7392ea4c,AudemardKR2020,Joao2020,darwiche2020reasons}, in particular decision tree models~\cite{izza2020explaining,HuangII021}, has been extensively explored in XAI.
Specifically,~\cite{QUINLAN1987221} demonstrated that small decision trees are more interpretable than larger ones, and \cite{HUYSMANS2011141} ran a user study to illustrate that larger representations result in a decrease in both the accuracy of user's answers and user’s confidence in the model itself.
This work supports the motivation that minimizing explanations is a valuable direction to explore.

\para{Ensemble Methods.}
The Cascading Decision Trees algorithm pursues a different goal than existing ensemble methods, such as Bagging~\cite{bagging96} and Boosting~\cite{Schapire1990}.
Ensemble methods combine multiple weaker classifiers into a larger classifier to increase overall accuracy, while Cascading Decision Trees focus on shortening explanations.
The combination of ensemble methods obfuscate the traceability of a classic decision tree, which is the source of simple explanations for classifications~\cite{constrained04zhao}.

\para{Oblique Decision Trees.}
Oblique decision trees~\cite{murthy94oblique} extend the classic decision tree model by allowing each decision node to combine checks against multiple features.
This series of work contains a rich family of models such as multivariate decision trees~\cite{multivariate95brodley}, loose and tight coupling~\cite{cascade00}, and constrained cascade generalization of decision trees~\cite{constrained04zhao}.
However, compared to Cascading Decision Trees, this oblique design is not only expensive in the training stage~\cite{Guang20oblique}, but it also obscures the explanation of a classification in the decision-making stage~\cite{constrained04zhao}.

\para{Optimal Decision Trees.}
Optimal decision trees learn models with the optimal prediction accuracy under different constraints, such as a certain number of leaves in the tree~\cite{sparse19} or restriction to complete binary trees of a certain depth~\cite{VerwerZ19}.
However, learning an optimal decision tree without heuristics is an NP-complete problem~\cite{constructing76Laurent}. 
The main drawback of using this approach in practice is that the model is too computationally expensive to train.
For example, state-of-the-art optimal decision trees implementations~\cite{AglinNS20,VerwerZ19} need ten minutes to train the ionosphere dataset\footnote{https://archive.ics.uci.edu/ml/datasets/Ionosphere}. 
On the same dataset, we show in our evaluation that our Cascading Decision Trees learning process terminated in seconds with competitive accuracy (cf.~\cite{VerwerZ19}). 
In addition, instead of being a direct competitor, an optimal decision tree learner could be used as the base decision tree inducer in our algorithm to potentially improve the prediction accuracy, though at the cost of training time.

\section{Motivating Examples}
\label{sec:motivation}

\setlength{\tabcolsep}{1em}
\begin{table}[t]
\centering
\caption{Synthetic Dataset for Binary Classification.}
\label{table:synthetic}
\begin{footnotesize}
\resizebox{1.0\columnwidth}{!}{
\begin{tabular}{|c|c|c|c|c|c|}
\hline
{ } & {\bf Feature1} & {\bf Feature2} & {\bf Feature3} & {\bf Feature4}  & {\bf Label}\\ \hline
Sample1     & T & T & T & F      & Positive     \\ \hline
Sample2     & T & T & F & F      & Positive     \\ \hline
Sample3     & T & T & T & F      & Positive     \\ \hline
Sample4     & T & T & T & F      & Positive     \\ \hline
Sample5     & F & F & F & T      & Positive     \\ \hline
Sample6     & F & T & F & F      & Positive     \\ \hline
Sample7     & T & F & F & F      & Negative     \\ \hline
Sample8     & F & T & T & F      & Negative     \\ \hline
Sample9     & F & F & T & F      & Negative     \\ \hline
Sample10    & F & T & F & F      & Negative     \\ \hline

\end{tabular}
}
\end{footnotesize}
\end{table}

\begin{figure}[t]
 \centering
      \begin{tikzpicture}[thick,scale=0.7, every node/.style={scale=0.45}]

\tikzstyle{block} = [node distance=0.6cm, align=center, inner sep=0.2cm]
\tikzstyle{decision} = [block, rectangle, draw, minimum width=1.5cm, minimum height=2em, thick]
\tikzstyle{leaf1} = [block, fill=green!10, rounded corners=5pt, align=center]
\tikzstyle{leaf2} = [block, fill=red!10, rounded corners=5pt, align=center]
\tikzstyle{leaf3} = [block, fill=gray!30, rounded corners=5pt, align=center]

    \node [decision] (root) {(1) Feature1};
    \node [decision, below left=of root] (rnd1) {(2) Feature2};
    \node [decision, below right=of root] (rnd2) {(3) Feature3};
    \node [decision, below right=of rnd2] (rnd3) {(4) Feature2};
    \node [leaf1, below left=of rnd1] (node1) {(5) S1,S2,S3,S4};
    \node [leaf2, right=of node1] (node2) {(6) S7};
    \node [leaf2, below left=of rnd2] (node3) {(7) S8,S9};
    \node [leaf2, below left=of rnd3] (node5) {(8) S6,S10};
    \node [leaf1, below right=of rnd3] (node4) {(9) S5};

    \draw [->] (root) -- node [above left]{True} (rnd1);
    \draw [->] (root) -- node [above right]{False} (rnd2);
    \draw [->] (rnd1) -- node [above left]{True} (node1);
    \draw [->] (rnd1) -- node [right]{False} ([xshift=0.08cm] node2.north);
    \draw [->] (rnd2) -- node [above left]{True} (node3);
    \draw [->] (rnd2) -- node [above right]{False} (rnd3);
    \draw [->] (rnd3) -- node [above left]{True} (node5);
    \draw [->] (rnd3) -- node [above right]{False} (node4);
\end{tikzpicture}
      \caption{Example of a full classic decision tree on dataset given in 
      Table~\ref{table:synthetic}, with green boxes represent Positive nodes and red boxes represent Negative nodes.}
      \label{fig:fulldtree}

\end{figure}
The following simple synthetic example is constructed
to illustrate the basic functionality of Cascading Decision Trees. 
Given the dataset in Table~\ref{table:synthetic}, a classic decision tree will construct a model as shown in Figure~\ref{fig:fulldtree}.
Using the same dataset, our Cascading Decision Trees algorithm generates a model with three subtrees as shown in Figure~\ref{fig:casdtree}.
Let us assume that there is a new sample ,\texttt{Sample11}, with the feature vector $(F, F, F, T)$.
Both models classify \texttt{Sample11} with the same prediction result, \texttt{Positive}.
However, the explanations extracted from these two models are different.

\begin{figure}[t]
 \centering
      \begin{tikzpicture}[thick,scale=0.45, every node/.style={scale=0.52}]

\tikzstyle{block} = [node distance=0.6cm, align=center, inner sep=0.2cm]
\tikzstyle{decision} = [block, rectangle, draw, minimum width=1.5cm, minimum height=2em, thick]
\tikzstyle{leaf1} = [block, fill=green!10, rounded corners=5pt, align=center]
\tikzstyle{leaf2} = [block, fill=red!10, rounded corners=5pt, align=center]
\tikzstyle{leaf3} = [block, fill=gray!30, rounded corners=5pt, align=center]

    \node [decision] (root) {(1) Feature1};
    \node [decision, below left=of root] (rnd1) {(2) Feature2};
    \node [decision, below right=of root] (rnd2) {(3) Feature3};
    \node [leaf1, below left=of rnd1] (node1) {(4) S1,S2,S3,S4};
    \node [leaf2, right=of node1] (node2) {(5) S7};
    \node [leaf2, below left=of rnd2] (node3) {(6) S8,S9};
    \node [leaf2, below right=of rnd2] (node4) {(7) S5,S6,S10};

    \draw [->] (root) -- node [above left]{True} (rnd1);
    \draw [->] (root) -- node [above right]{False} (rnd2);
    \draw [->] (rnd1) -- node [above left]{True} (node1);
    \draw [->] (rnd1) -- node [right]{False} ([xshift=0.145cm] node2.north);
    \draw [->] (rnd2) -- node [above left]{True} (node3);
    \draw [->] (rnd2) -- node [above right]{False} (node4);

    \node [decision, below=2.6cm of root] (root_) {(8) Feature4};
    \node [decision, below right=of root_] (rnd1_) {(9) Feature3};
    \node [leaf1, below left=of root_] (node1_) {(10) S5};
    \node [leaf2, below left=of rnd1_] (node2_) {(11) S8,S9};
    \node [leaf2, below right=of rnd1_] (node3_) {(12) S6,S7,S10};

    \draw [->] (root_) -- node [above left]{True} (node1_);
    \draw [->] (root_) -- node [above right]{False} (rnd1_);
    \draw [->] (rnd1_) -- node [above left]{True} (node2_);
    \draw [->] (rnd1_) -- node [above right]{False} (node3_);

    \node [decision, below=2.6cm of root_] (root__) {(13) Feature3};
    \node [decision, below right=of root__] (rnd1__) {(14) Feature1};
    \node [leaf2, below left=of root__] (node1__) {(15) S8,S9};
    \node [leaf2, below left=of rnd1__] (node2__) {(16) S7};
    \node [leaf2, below right=of rnd1__] (node3__) {(17) S6,S10};

    \draw [->] (root__) -- node [above left]{True} (node1__);
    \draw [->] (root__) -- node [above right]{False} (rnd1__);
    \draw [->] (rnd1__) -- node [above left]{True} (node2__);
    \draw [->] (rnd1__) -- node [above right]{False} (node3__);

    \draw [->, dotted] (node2) |- (root_) node [midway,fill=white] {};
    \draw [->, dotted] ([xshift=0.8cm] node3) -- (root_) node [midway,fill=white] {};
    \draw [->, dotted] (node4) |- (root_) node [midway,fill=white] {};

    \draw [->, dotted] ([xshift=0.5cm] node2_) -- (root__) node [midway,fill=white] {};
    \draw [->, dotted] (node3_) |- (root__) node [midway,fill=white] {};

\end{tikzpicture}
      \caption{Example of Cascading Decision Trees on dataset given in Table\ref{table:synthetic}, with maximum depth of two in each subtree.}
      \label{fig:casdtree}
\end{figure}

In the classic decision tree model (Figure~\ref{fig:fulldtree}), \texttt{Sample11} falls into \texttt{node(9)}.
Thus, the explanation path is \texttt{Feature1 = F}, \texttt{Feature2 = F} and \texttt{Feature3 = F}, with 
an explanation length of three.

Using Cascading Decision Trees (Figure~\ref{fig:casdtree}), \texttt{Sample11} 
first falls into \texttt{node(7)} in the first subtree.
\texttt{node(7)} is not classifying positive samples, so \texttt{Sample11}  is passed to the second subtree and falls eventually into \texttt{node(10)}.
This way, the explanation path for \texttt{Sample11} is \texttt{Feature4 = T}, with an explanation length of only one.

\if
However, from this dataset, we observe that ``Feature4 = T'' alone is enough to classify ``Sample11'' as ``Positive''. \ruzica{Feature4 doesn't even appear in the decision tree, how can we then classify it as positive? Also, looking at the table, Feature4 = T only once, so how can we then conclude that if Feature4 = T, then it is a positive label. Very unconvincing!}
This explanation has depth only one and it is significantly shorter compared to the justification provided by the classic decision trees.
To have this explanation, instead of having one large tree to do classification, we have several cascading decision subtrees. \ruzica{not clear what are those. also this is not an insight, maybe ``we aim'' but i cannot see what is an insight here?}
Each one of decision subtree has its depth limit, and it explains sufficient reasons behind a ``Positive'' classification.

All samples already classified as ``Positive'' by the current subtree are removed from the training data in the next round decision tree construction. \ruzica{are you now describing the algorithm for constructing a cascading decision tree? if yes, why on earth the algorithm description in the motivation section? are you doing something else. please exaplain what are you doing because it is not clear}
For example, in Figure~\ref{fig:casdtree}, ``Sample1 - Sample4'' are removed from the training data since we already capture the explanation for these four ``Positive'' classifications.
Only ``Sample5 - Sample10'' are then passed into the second decision subtree construction.
We build the decision subtree iteratively until there are no more leaf nodes which are classified as ``Positive''.
In this example, we end up building our cascading decision trees with three subtrees.

To classify the ``Sample11'', we pass this sample to our cascading decision trees model in a sequence.
In the first subtree, this sample ends in the node (7), which is not classified as ``Positive'', so we pass this sample into our second subtree.
This sample falls into the node (10) in the second subtree in Figure\ref{fig:casdtree}.
Our model manages to report the succinct explanation ``Feature4 = T'' to users as explanations.

\ruzica{this section needs to be more to the point and more focuses. i already forgot what did you try to sell me. i try to understand what is happening, but there was no connection nor flow between the paragraphs}
\fi
\section{Invariant-based Explanation and Explanation Path}
\label{sec:explanation}

\subsection{Invariant-based Explanation}
For interpretability, we take inspiration from the LIME system~\cite{ribeiro2016should}, which states that interpretable explanations of a model should ``provide qualitative understanding between joint values of input variables and the resulting predicted response value''.
An explanation reasonably defined should capture the essence of the model's output.
That is, the explanation identifies the key input values that contributed to the output of the model.
The output of the model should be an invariant of the input values not included in the explanation.

Denote the input data by $x \in \mathcal{X}$ and the label of the class by $y \in \mathcal{Y}$.
Let $f(x;w):\mathcal{X}\rightarrow \mathcal{Y}$ be the model taking $x$ as input and parameterized by $w$.
Multiclass classification can be reduced to binary classification in natural ways.
For simplicity, we specifically focus on binary classification tasks in this work.
Hence we assume $\mathcal{Y} = \{0,1\}$.
Although there are various kinds of structures of $\mathcal{X}$, the most common one is $\mathcal{X} = \{0,1\}^d$.
Usually other typical features in $\mathcal{X}$ can be transformed to the form of $\{0,1\}^d$.
For example, categorical feature $x \in \{C_1,\cdots,C_m\}$ can be converted to $m$ dummy features $(z^{(1)},\cdots,z^{(m)})$ where $z^{(i)} = \mathbbm{1}[x = C_i]$.

For a model $f(x;w)$, we define a valid explanation $e^f(x;w)$ for $x$ as definition \ref{def:valid_explanation}.

\begin{definition}[Valid Explanation]\label{def:valid_explanation}
    Let $f(x;w)$ denote a model parameterized by $w$ and taking $x$ as input.
	An explanation for $x$ on model $f(x;w)$ is a vector $e^f(x;w)\in \{0,1\}^d$.
	An explanation $e^f(x;w)$ is called a valid explanation for $x$ on $f(x;w)$ if and only if for all $z\in\mathcal{X}$,
	\begin{align*}
		f(x;w) = f\left((x* e^f(x;w)) \bigoplus (z* \neg e^f(x;w)); w\right)
	\end{align*}
	where $*$ is the element-wise logical AND operator, $\bigoplus$ is the element-wise logical XOR operator, and $\neg$ is the element-wise NOT operator.
\end{definition}

Lemma \ref{lem:invariant} shows why our definition of valid explanation is reasonable in the aspects of capturing the values in the input contributing to the model's output.
If feature $i$ is not picked by the explanation, i.e. $e^f(x;w)_i = 0$, then any disturbance on feature $i$ would not change the model's output.
\begin{lemma}[Invariant]\label{lem:invariant}
	Suppose $x$ is a given input, $f(x;w)$ is a given model, and $e^f(x;w)$ is a valid explanation.
	If $x_i = y_i$ for all $i$ such that $e^f(x;w)_i = 1$, then $f(y;w) = f(x;w)$.
\end{lemma}

\begin{proof}
	Suppose $x=(x_1,\cdots,x_d)$ and $y=(y_1,\cdots,y_d)$ satisfying $x_i = y_i$ for all $i\in I$ where $I$ is defined as $I:=\{i: e^f(x;w)_i = 1\}$.
	Construct $z\in\mathcal{X}$ as
	\begin{equation*}
		z_i = \left\{
		\begin{aligned}
		 & 0 & ~~~ e^f(x;w)_i = 1 \\
		 & y_i & ~~~ e^f(x;w)_i = 0
		\end{aligned}
		\right.
	\end{equation*}
	Thus we can verify that
	\begin{align*}
		y = (x* e^f(x;w)) \bigoplus (z* \neg e^f(x;w))
	\end{align*}
	Since $e^f$ is a valid explanation, $f(y;w) = f(x;w)$ holds.
\end{proof}

A well-known principle named Occam's razor posited the philosophy that shorter, simpler explanations are more likely to be true. Therefore, we define the length of a valid explanation by definition
\ref{def:length_of_explanation}.

\begin{definition}[Length of Valid Explanation]\label{def:length_of_explanation}
	Suppose that $e^f(x;w)$ is a valid explanation, the length of $e^f(x;w)$ is
	\begin{align*}
		|e^f(x;w)| = \sum_{i=1}^d e^f(x;w)_i
	\end{align*}
\end{definition}

We provide the following examples to further illustrate our definition of a valid explanation.

\textbf{Example 1}
A trivial explanation is always available for any sample on any model - the explanation that every element in the input vector is important.
Let $f(\cdot;w)$ be the decision tree described by Figure~\ref{fig:fulldtree}, and let input $x$ be \texttt{Sample5}.
Explanation $e^f(x;w) = (1,1,1,1)$ is a valid explanation.
Notice that $\{i:e^f(x;w)_i = 0\} = \emptyset$, which means that as long as no features change, the output of the model remains the same.
In this example, $|e^f(x;w)| = 4$.

\textbf{Example 2} 
In general, a shorter explanation can be extracted from classic decision trees.
As an example, the explanation for $x=$\texttt{Sample5} on the decision tree in Figure~\ref{fig:fulldtree} is the explanation that keeps only the elements in the decision path of the tree: $e^f(x;w) = (1,1,1,0)$.
This is the commonly accepted approach to explanations of decision tree classifications.
In this example, $|e^f(x;w)| = 3$.

\textbf{Example 3} 
An even shorter explanation is possible with Cascading Decision Trees.
Looking to $x=$\texttt{Sample5} as classified by the Cascading Decision Trees in Figure~\ref{fig:casdtree}, a valid explanation keeps only \texttt{Feature 4}: $e^f(x;w) = (0,0,0,1)$. 
Although changes to \texttt{Feature 1-3} may change the decision path, the output remains the same as long as \texttt{Feature 4} is unchanged.
In this example, $|e^f(x;w)| = 1$.

\begin{definition}[Valid Explanatory Model]
	The pair of $M = (f,e^f)$ is called an explanatory model, where $f$ is the original model and $e^f$ is the associated explanation.
	An explanatory model $M = (f,e^f)$ is valid if and only if $e^f(x;w)$ is valid for all $x\in\mathcal{X}$.
\end{definition}


\subsection{Decision Path and Explanation Path in Decision Tree Models}
In this work, we represent a decision tree model $T$ by its nodes $V_T = \{\sigma_j\}\cup \{\tau_k\}$.
Denote non-leaf nodes by $\sigma_j = (i_j, l_j, r_j) \in [d]\times V_T \times V_T$, where $i_j$ means node $\sigma_j$ splits into two branches based on value of feature $x_{i_j}$.
The model goes to node $l_j$ if $x_{i_j} = 1$ and goes to node $r_j$ if $x_{i_j} = 0$.
Denote leaf nodes by $\tau_k \in \{0,1\}$.
The model outputs $f(x;w) = \tau_k$ if it finishes at leaf node $\tau_k$.

We define decision path for an input on a model as all the nodes visited by the model in definition \ref{def:decision_path}.
\begin{definition}[Decision Path]\label{def:decision_path}
	Let $x\in\mathcal{X}$ be an input and $f(x;w)$ be a decision tree model. Suppose in the computation of $f(x;w)$ the model goes through nodes $\sigma_{j_1},\cdots,\sigma_{j_m},\tau_k$ sequentially, i.e. for $t\in[m-1]$
	\begin{equation*}
		\sigma_{j_{t+1}} = \left\{
			\begin{aligned}
				l_{j_{t}} & ~~~ \text{if } x_{i_{j_{t}}} = 1 \\
				r_{j_{t}} & ~~~ \text{if } x_{i_{j_{t}}} = 0
			\end{aligned}
		\right.
		~ \text{ and } ~ \tau_k = \left \{
			\begin{aligned}
				l_{j_{m}} & ~~~ \text{if } x_{i_{j_{m}}} = 1 \\
				r_{j_{m}} & ~~~ \text{if } x_{i_{j_{m}}} = 0
			\end{aligned}
		\right.
	\end{equation*}
	then the decision path for $x$ on $f$ is $\langle \sigma_{j_1},\cdots,\sigma_{j_m} \rangle$.
\end{definition}

The explanation path is defined as a path that can be used to generate a valid explanation.
\begin{definition}[Explanation Path]\label{def:explanation_path}
	Let $x \in \mathcal{X}$ be an input and $f(x;w)$ be a decision tree model. A path $p = \langle \sigma_{j_1}, \cdots, \sigma_{j_m} \rangle$ is called an explanation path for $x$ on $f$, if the explanation $e$ is a valid explanation for $x$ on $f$, where for $s\in[d]$,
	\begin{align*}
		e^f(x;w)_s = \mathbbm{1}[\exists \sigma_t \in p, i_{t} = s]
	\end{align*}
\end{definition}

Notice that for an input $x$ and a decision tree model $f(x;w)$, the output is always determined by the values of features on the decision path. Hence the decision path is always an explanation path as shown in lemma \ref{lem:decision_path_is_explanation_path}.
\begin{lemma}\label{lem:decision_path_is_explanation_path}
	Let $x \in \mathcal{X}$ be an input and $f(x;w)$ be a decision tree model. Suppose $p = \langle \sigma_{j_1}, \cdots, \sigma_{j_m} \rangle$ be the decision path for $x$ on $f$. Then $p$ is an explanation path for $x$ on $f$.
\end{lemma}
\begin{proof}
   Let $e$ be the explanation for $x$ on $f$ generated by $p = \langle \sigma_{j_1}, \cdots, \sigma_{j_m} \rangle$ as
   \begin{align*}
      e^f(x;w)_s = \mathbbm{1}[\exists \sigma_t \in p, i_{t} = s]
   \end{align*}
   For any $z \in \mathcal{X}$, let $y = (x* e) \bigoplus (z* \neg e)$.
   Consider the decision path $q = \langle \sigma_{j'_1},\cdots,\sigma_{j'_{m'}} \rangle$ for $y$ on $f$.
   Notice that the first node in the decision path for any input on $f$ should be the root node of the decision tree.
   Therefore we have $\sigma_{j'_1} = \sigma_{j_1}$.
   Also notice that if $\sigma_{j'_t} = \sigma_{j_t}$, then $\sigma_{j'_{t+1}} = \sigma_{j_{t+1}}$ because $y_{i'_{t+1}} = x_{i_{t+1}}$.
   Therefore by mathematical induction, we know $q = p$.
   Since the decision path for $y$ on $f$ is the same as the decision path for $x$ on $f$, we have $f(y;w) = f(x;w)$.
   Thus $e$ generated by path $p$ is a valid explanation for $x$ on $f$, which by definition \ref{def:explanation_path} means $p$ is an explanation path.
\end{proof}

\section{Cascading Decision Trees}
\label{sec:method}

We introduce Cascading Decision Trees to show we can achieve a shorter explanation path by separating the decision path and explanation path on a decision tree model.
Notice that cascading decision trees to generate shorter explanation path for negative samples can be built symmetrically as building the cascading decision trees for positive samples.
So for simplicity, we only introduce the algorithm for building cascading decision trees for positive samples in this section.

\para{Building Cascading Decision Trees.}
The cascading decision trees algorithm is described through pseudocode in Algorithm~\ref{algo:train} (See in the supplementary material).  
Our insight is in the training process - it is the goal of each cascading decision subtree to identify the smallest subset of features that can classify as many \texttt{Positive} samples as possible, without misclassifying \textit{any} \texttt{Negative} samples. 
Specifically, the procedure \text{Fit} builds a classic decision tree, \texttt{clf}, on our training set, $\trainingSet$ and adds it to the cascading tree list - initially this list is empty.
For every leaf node we compute the \texttt{mixed} value, which is the percentage of samples from $\trainingSet$ classified by this leaf node that are also \texttt{Positive} samples.
\texttt{Positive} nodes are leaf nodes with a \texttt{mixed} value greater than the threshold.
If \texttt{clf} has no \texttt{Positive} nodes, it means we have learned a sufficiently good classifier and we stop (Line 11).
Otherwise, we first remove samples truly classified by \texttt{Positive} nodes in this \texttt{clf} from \trainingSet, and then obtain a new \trainingSet to use in the next iteration of the loop (Line 14).

Our cascading decision trees Algorithm~\ref{algo:train} presented here has no pruning phase. 
However, our algorithm is generic enough to be combined with any pruning techniques~\cite{esposito97pruning} and different goodness measurement for decision nodes split, such as entropy~\cite{entropyShannon} and gini impurity~\cite{Havrda67quantificationmethod}.

The time complexity of cascading decision tree algorithm is bounded by the size of the training set.
Suppose the training time for building classic decision trees is a function of the number of the training samples, $\ensuremath{\mathnormal{n}}$ and the number of features $\ensuremath{\mathnormal{d}}$.
We use the decision tree module in scikit-learn~\cite{scikit-learn} as our base classic decision tree inducer, which is built upon the CART~\cite{Breiman84Classification} method.
Since features are recursively reused in every decision node based on a numerical splitting criterion, the depth of the decision tree is bounded by the number of the training samples $\ensuremath{\mathnormal{n}}$ in our model.
Therefore, the time complexity for building one base decision tree is bounded by $\mathcal{O}(n^2d)$.

According to our cascading decision trees Algorithm~\ref{algo:train}, after building one decision subtree, samples that are classified as \texttt{True Positive} are removed from the next round of decision subtree construction.
In the worst case, every time, only one \texttt{True Positive} is classified in the current decision tree. 
The time complexity for building the next cascading decision subtree is in 
$\mathcal{O}(n^2d)$.
Therefore, the overall cascading decision trees training time $\ensuremath{\mathnormal{T}}$ is bounded by:
$\mathcal{O}(\sum_{m=1}^{n}m^2d) = \mathcal{O}(\frac{n(n+1)(2n+1)}{6}d) = \mathcal{O}(n^3d)$.

\para{Cascading Decision Trees Inference.} Our cascading decision trees testing process is described in Algorithm~\ref{algo:test} in the supplementary material.
In the procedure \text{Test}, we run all decision trees sequentially, and we report the decision path of only the classifying subtree as our explanation path.
We also take the missing value into consideration.
If in current decision tree we meet any missing value, we will skip the current decision tree and goes to the next decision tree.

%
%

\section{Experiments}
\label{sec:eval}
This section aims to evaluate the Cascading Decision Trees algorithm by answering the following questions:
\begin{enumerate}
\item How does the explanation length for Cascading Decision Trees compare to the baselines?
\item What is the prediction accuracy and efficiency of the Cascading Decision Trees algorithm?
\item How does the Cascading Decision Trees algorithm perform against missing data?
\item How does the Cascading Decision Trees algorithm perform in real-world application such as continuous integration (CI) build status prediction?
\end{enumerate}

\para{Experiment Setup.} To evaluate Cascading Decision Trees, we collect ten binary classification datasets from UCI machine learning repository\footnote{http://archive.ics.uci.edu/ml/index.php/}.
These datasets are standard benchmarks for comparing the performance of tree-based classification methods~\cite{constrained04zhao}.
Additionally, since the underlying library, scikit-learn\footnote{For a fair comparison, the decision trees module in scikit-learn is also used for all three baselines in our evaluation.}, of our implementation needs numerical features, Cascading Decision Trees adopts a standard one-hot encoder to add several indicator variables in the datasets. Explanations are computed using the features after encoding.

\para{Baselines.} We compared the accuracy of our model against three baselines.
They are 1) classic decision trees with PI-explanation, 2) sequential covering, and 3) classic decision trees with bounded depth.
They were designed to provide better interpretability than the classic decision trees model while still preserving the prediction accuracy.

Decision trees with prime implicant (PI) explanations~\cite{ShihCD18,NEURIPS2019_7392ea4c,AudemardKR2020,Joao2020,izza2020explaining,darwiche2020reasons} serve as our major baseline in this paper, since they target providing the minimal explanation to the decision tree model.
PI-explanation is computed in the post-hoc fashion (decision rule pruning) for instances classified by the model~\cite{ShihCD18,izza2020explaining}.
It first builds a classic decision tree model and then computes a subset-minimal set of feature values that entail the same prediction.
Thus it provides a explanation no longer than the original, unpruned decision tree model without any loss of prediction accuracy.
We denote PI-explanation as the name of this baseline in our paper.

Sequential covering~\cite{COHEN1995115,Furnkranz99} is a popular, separate-and-conquer rule learning algorithm based on the closed-world assumption.
Every time, it learns a single rule, removes the data that this rule covers and then repeats the same process until the terminated condition is reached.
It finally connects all of the learned rules sequentially during the training phase. 
Sequential covering has many variants, in our setting, every time, the algorithm tries to learn a rule that classifies as many positive examples as possible while keeping the negative examples not covered.

Classic decision trees with bounded depth uses the library in scikit-learn as our base classic decision tree inducer, which is built upon the CART~\cite{Breiman84Classification} method.

We quantify algorithm performance using five-fold cross validation, and randomly shuffle the datasets.
The maximum depth of the Cascading Decision Trees algorithm is uniformly set to two in all tests.
The $\theta$ in Cascading Decision Trees algorithm is set to vary from 0.5 to 0.9 based on the percentage of positive samples in the dataset.
We ran all experiments on a Macbook Pro with an Inter i7 CPU and 16GB of RAM.

\para{Explainability.}
The shorter an explanation, the more comprehensible that explanation is to users~\cite{QUINLAN1987221,HUYSMANS2011141,knowledge00Pazzani}.
Larger representations result in a decrease in both user's answer accuracy and confidence in the model itself~\cite{HUYSMANS2011141}.

\bgroup
\def\arraystretch{1.15}
\begin{table}[t]
\centering
\caption{Average explanation length of positive classifications from PI-explanation and Cascading Decision Trees.}
\label{table:depth_comparison}
\begin{footnotesize}
\resizebox{\columnwidth}{!}{
\begin{tabular}{|l|l|l|l|}
\hline
Dataset             & PI-explanation& Cascading Decision Trees & Improvement  \\ \hline
Breast cancer  		&  7.854 &  2.000 & 74.5\%     \\ \hline
Heart Disease 		&  4.377 &  2.000 & 54.3\%     \\ \hline
Wisc Breast cancer  &  2.658 &  1.991 & 25.1\%     \\ \hline
Cervical Cancer  	&  7.408 &  2.000 & 73.0\%     \\ \hline
Credit Card			&  6.408 &  2.000 & 68.8\% 	   \\ \hline
German Credit		&  9.788 &  2.000 & 79.6\% 	   \\ \hline
Climate				&  2.945 &  2.000 & 32.1\% 	   \\ \hline
Happiness			&  7.315 &  2.000 & 72.7\% 	   \\ \hline
Ionosphere    		&  2.694 &  1.418 & 47.4\%     \\ \hline
Sonar         		&  3.813 &  1.943 & 49.0\%     \\ \hline

\end{tabular}
}
\end{footnotesize}
\end{table}
\egroup

\bgroup
\def\arraystretch{1.22}
\begin{table*}[t]
\centering
\caption{Breakdown Evaluation of Cascading Decision Trees on Ten UCI datasets.}
\label{table:Result}
\begin{footnotesize}
\resizebox{2\columnwidth}{!}{
\begin{tabular}{|l|l|l|l|l|l|l|l|l|l|l|l|}
\hline
Dataset                            & Method 				    & \vtop{\hbox{\strut Explanation}\hbox{\strut Length}}  & Accuracy & Runtime (s) & TP 	  & TN    &  FP   & FN    & Precision & Recall    & F-1 Score \\ \hline
\multirow{4}{*}{Breast Cancer}      & Cascading Decision Trees  				& \bf{2.000}	     & 73.10\%  & 0.392  	  & 4.4   & 38.0  &  2.6  &  13.0 & 67.56\%   &  26.05\%  & 36.33\%   \\ \cline{2-12} 
									& PI-explanation       		& 7.854	  			 & 64.48\%  & 220.0 	  & 7.4   & 30.0  &  10.6 &  10.0 & 40.74\%   &  43.22\%  & 40.96\%   \\ \cline{2-12}
									& Sequential Covering		& 4.888	  			 & 63.45\%  & 0.987 	  & 8.8   & 28.0  &  12.6 &  8.6  & 44.24\%   &  48.16\%  & 41.83\%   \\ \cline{2-12}
									& Classic (max\_depth = 3)  & 3.000	  			 & 71.03\%  & 0.065 	  & 3.6   & 37.6  &  3.0  &  13.8 & 64.89\%   &  21.72\%  & 29.59\%   \\ \hline

\multirow{4}{*}{Heart Disease}      & Cascading Decision Trees        		& \bf{2.000}	     & 74.07\%  & 0.428  	  & 12.4  & 27.6  &  2.4  &  11.6 & 87.76\%   &  51.22\%  & 62.20\%   \\ \cline{2-12} 
									& PI-explanation       		& 4.377	  			 & 72.96\%  & 57.40 	  & 15.8  & 23.6  &  6.4  &  8.2  & 70.62\%   &  66.19\%  & 67.71\%   \\ \cline{2-12}
									& Sequential Covering		& 3.465				 & 75.19\%  & 0.737	      & 16.2  & 24.4  &  5.6  &  7.8  & 74.17\%   &  68.58\%  & 70.92\%   \\ \cline{2-12}
									& Classic (max\_depth = 3)  & 3.000	  			 & 74.07\%  & 0.061 	  & 12.8  & 27.2  &  2.8  &  11.2 & 84.70\%   &  51.10\%  & 60.97\%   \\ \hline

\multirow{4}{*}{Wisc Breast Cancer} & Cascading Decision Trees        		& \bf{1.991}    	 & 93.51\%  & 1.003  	  & 36.8  & 69.8  &  1.8  &  5.6  & 95.42\%   &  86.57\%  & 90.54\%   \\ \cline{2-12} 
                                    & PI-explanation       		& 2.658    			 & 93.16\%  & 26.17       & 38.4  & 67.8  &  3.8  &  4.0  & 91.38\%   &  90.47\%  & 90.80\%   \\ \cline{2-12}
                                    & Sequential Covering		& 2.377				 & 93.68\%  & 1.018		  & 39.4  & 67.4  &  4.2  &  3.0  & 90.25\%   &  92.94\%  & 91.51\%   \\ \cline{2-12}
                                    & Classic (max\_depth = 3)  & 2.311	  			 & 91.40\%  & 0.077  	  & 34.8  & 69.4  &  2.2  &  7.6  & 93.88\%   &  82.07\%  & 87.53\%   \\ \hline

\multirow{4}{*}{Cervical Cancer}    & Cascading Decision Trees  				& \bf{2.000}	     & 96.12\%  & 1.013  	  & 8.0   & 120.8  &  4.2  &  1.0 & 67.18\%   &  87.07\%  & 75.04\%   \\ \cline{2-12} 
									& PI-explanation       		& 7.048	  			 & 94.63\%  & 25.43 	  & 4.8   & 122.0  &  3.0  &  4.2 & 64.85\%   &  51.80\%  & 56.49\%   \\ \cline{2-12}
									& Sequential Covering		& 5.383	  			 & 95.37\%  & 0.987 	  & 4.6   & 123.2  &  1.8  &  4.4 & 71.67\%   &  50.02\%  & 58.85\%   \\ \cline{2-12}
									& Classic (max\_depth = 3)  & 3.000	  			 & 95.52\%  & 0.101 	  & 6.6   & 121.4  &  3.6  &  2.4 & 64.59\%   &  71.51\%  & 63.72\%   \\ \hline

\multirow{4}{*}{Credit Card}        & Cascading Decision Trees        		& \bf{2.000}	     & 86.41\%  & 0.743  	  & 57.6  & 55.6  &  3.6  &  14.2 & 94.15\%   &  80.30\%  & 86.67\%   \\ \cline{2-12} 
									& PI-explanation       	   	& 6.408	  			 & 80.61\%  & 490.8 	  & 59.4  & 46.2  &  13.0 &  12.4 & 82.29\%   &  82.76\%  & 82.37\%   \\ \cline{2-12}
									& Sequential Covering		& 4.656				 & 81.07\%	& 1.475		  & 60.0  & 46.2  &  13.0 &  11.8 & 84.53\%   &  83.34\%  & 82.72\%   \\ \cline{2-12}
									& Classic (max\_depth = 3)  & 3.000	  			 & 85.95\%  & 0.086 	  & 57.0  & 55.6  &  3.6  &  14.8 & 94.11\%   &  79.37\%  & 86.10\%   \\ \hline

\multirow{4}{*}{German Credit}      & Cascading Decision Trees        		& \bf{2.000}	     & 71.80\%  & 1.573  	  & 20.4  & 123.2 &  16.8 &  39.6 & 50.97\%   &  33.23\%  & 38.84\%   \\ \cline{2-12} 
									& PI-explanation       		& 9.788	  			 & 66.90\%  & 3173.4 	  & 28.2  & 105.6 &  34.4 &  31.8 & 45.06\%   &  47.10\%  & 45.93\%   \\ \cline{2-12}
									& Sequential Covering   	& 4.774	  			 & 68.40\%  & 3.304 	  & 22.2  & 114.6 &  25.4 &  37.8 & 53.60\%   &  37.88\%  & 39.24\%   \\ \cline{2-12}
									& Classic (max\_depth = 3)	& 3.000	  			 & 73.00\%  & 0.143 	  & 22.2  & 123.8 &  16.2 &  37.8 & 56.93\%   &  36.75\%  & 43.41\%   \\ \hline

\multirow{4}{*}{Climate}            & Cascading Decision Trees        		& \bf{2.000}	     & 94.81\%  & 1.640  	  & 96.8  & 5.6 &  3.6 &  2.0 & 96.41\%   &  97.98\%  & 97.18\%   \\ \cline{2-12} 
									& PI-explanation       		& 2.945	  			 & 93.15\%  & 77.14 	  & 95.0  & 5.6 &  3.6 &  3.8 & 96.34\%   &  96.17\%  & 96.25\%   \\ \cline{2-12}
									& Sequential Covering   	& 2.455	  			 & 90.74\%  & 1.070 	  & 93.2  & 4.8 &  4.4 &  5.6 & 95.48\%   &  94.31\%  & 94.88\%   \\ \cline{2-12}
									& Classic (max\_depth = 3)	& 2.541	  			 & 91.11\%  & 0.090 	  & 92.4  & 6.0 &  3.2 &  6.4 & 96.64\%   &  93.51\%  & 95.04\%   \\ \hline

\multirow{4}{*}{Happiness}			& Cascading Decision Trees        		& \bf{2.000}	     & 61.37\%  & 0.262  	  & 6.8  & 11.0 &  2.4 &  8.8 & 72.71\%   &  44.01\%  & 52.83\%   \\ \cline{2-12} 
									& PI-explanation       		& 7.315	  			 & 56.55\%  & 97.71 	  & 9.2  & 7.2  &  6.2 &  6.4 & 60.74\%   &  59.03\%  & 59.03\%   \\ \cline{2-12}
									& Sequential Covering   	& 5.178	  			 & 57.93\%  & 0.607 	  & 10.8  & 6.0 &  7.4 &  4.8 & 64.42\%   &  69.02\%  & 62.25\%   \\ \cline{2-12}
									& Classic (max\_depth = 3)	& 3.000	  			 & 55.86\%  & 0.053 	  & 6.4  & 9.8  &  3.6 &  9.2 & 60.79\%   &  41.41\%  & 46.89\%   \\ \hline						

\multirow{4}{*}{Ionosphere}         & Cascading Decision Trees        		& \bf{1.418}    	 & 88.73\%  & 1.063 	  & 20.4  & 42.6  &  3.0  &  5.0  & 88.28\%   &  80.89\%  & 84.37\%   \\ \cline{2-12} 
                                    & PI-explanation       		& 2.694   			 & 85.92\%  & 14.09 	  & 20.0  & 41.0  &  4.6  &  5.4  & 81.77\%   &  80.16\%  & 80.68\%   \\ \cline{2-12}
									& Sequential Covering		& 1.681				 & 88.45\%  & 0.827		  & 21.0  & 41.8  &  3.8  &  4.4  & 85.89\%   &  83.06\%  & 84.16\%   \\ \cline{2-12}
									& Classic (max\_depth = 3)  & 1.483    			 & 84.23\%  & 0.072  	  & 15.8  & 44.0  &  1.6  &  9.6  & 92.42\%   &  61.57\%  & 73.60\%   \\ \hline

\multirow{4}{*}{Sonar}              & Cascading Decision Trees        		& \bf{1.943}	     & 66.19\%  & 0.468  	  & 12.2  & 15.6  &  4.2  &  10.0 & 78.63\%   &  55.92\%  & 62.46\%   \\ \cline{2-12} 
                                    & PI-explanation       		& 3.813	  			 & 74.29\%  & 23.37 	  & 17.4  & 13.8  &  6.0  &  4.8  & 74.45\%   &  78.29\%  & 76.06\%   \\ \cline{2-12}
                                    & Sequential Covering		& 3.019				 & 74.29\%  & 0.721       & 17.4  & 13.8  &  6.0  &  4.8  & 75.28\%   &  78.89\%  & 76.29\%   \\ \cline{2-12} 
                                    & Classic (max\_depth = 3)  & 2.658	  			 & 65.71\%  & 0.071 	  & 11.0  & 16.6  &  3.2  &  11.2 & 79.75\%   &  50.42\%  & 60.32\%   \\ \hline

\end{tabular}}
\end{footnotesize}
\end{table*}
\egroup


    \pgfplotsset{
        compat=1.9,
        %
        blank pyramid axis style/.style={
            width=0.3*\textwidth,
            height=0.3*\textheight,
            scale only axis,
            xmin=0,
            xmax=100,
            ymin=-0.5,
            ymax=9,
            y dir=reverse,
            enlarge y limits={value=0.075,upper},
            xbar,
            axis x line=left,
            xtick align=outside,
            bar width=1,
            allow reversal of rel axis cs=false,
        },
        pyramid axis style/.style={
            blank pyramid axis style,
            %
            xticklabel={%
                \pgfmathprintnumber\tick\%%
            },
            ytick=\empty,
            axis line style={-},
            %
            nodes near coords={%
                \pgfmathprintnumber\pgfplotspointmeta\%%
            },
            every node near coord/.append style={
                font=\small,
                color=black,
                /pgf/number format/fixed,
            },
        },
    }
\begin{figure}[t]
\centering
    \begin{tikzpicture}[scale=0.60]
        \pgfplotstableread[
            col sep=comma,
            header=true,
        ]{
            distribution,cascading,classic
            +10,0,20
            9--10,0,2
            8--9,0,10
            7--8,0,7
            6--7,0,16
            5--6,0,12
            4--5,0,6
            3--4,0,11
            2--3,0,6
            1--2,35,0
        }\loadedtable
        \pgfplotstablecreatecol[
            expr accum={
                round(\pgfmathaccuma) + \thisrow{cascading} + \thisrow{classic}
            }{0}
        ]{sum}{\loadedtable}
        \tikzset{
            fpu=true,
        }
            \pgfplotstablegetrowsof{\loadedtable}
                \pgfmathsetmacro{\LastRow}{\pgfplotsretval-1}
            \pgfplotstablegetelem{9}{sum}\of{\loadedtable}
                \pgfmathsetmacro{\Sum}{\pgfplotsretval}
        \tikzset{
            fpu=false,
        }
        \begin{axis}[
            pyramid axis style,
            %
            axis y line*=left,
            ytick=\empty,
            name=popaxis,
        ]
            \addplot [cyan,fill=cyan!70] table [
                y expr =\coordindex,x expr={\thisrow{classic}/90*100},
            ] \loadedtable;

            \node [anchor=south] at (rel axis cs:0.65,0.6)
                {\textcolor{cyan}{PI-explanation}};
        \end{axis}

        \begin{axis}[
            pyramid axis style,
            %
            at={(popaxis.west)},
            anchor=east,
            xshift=-12.5mm,
            %
            x dir=reverse,
            every node near coord/.append style={
                anchor=east,
            },
            axis y line*=right,
        ]
            \addplot [teal,fill=teal!65] table [
                y expr =\coordindex, x expr={\thisrow{cascading}/\thisrow{cascading}*100},
            ] \loadedtable;

            \node [anchor=south] at (rel axis cs:0.6,0.6)
                {\textcolor{teal}{Cascading Decision Trees}};
        \end{axis}

        \begin{axis}[
            blank pyramid axis style,
            %
            at={(popaxis.west)},
            anchor=east,
            xshift=-12.5mm,
            %
            x dir=reverse,
            axis y line*=right,
            xtick=\empty,
            ytick=data,
            yticklabels from table={\loadedtable}{distribution},
            y tick label style={
                align=center,
                inner sep=0pt,
                text width=12.5mm,
            },
            major tick length=0pt,
            axis line style={
                -,
                draw=none,
            },
        ]
            \addplot [draw=none,fill=none] table [
                y expr =\coordindex, x expr={0},
            ] \loadedtable;

        \end{axis}
    \end{tikzpicture}
\caption{A distribution of explanation lengths for all the examples in the Breast Cancer dataset. The explanation length for Cascading Decision Trees model is no greater than two, while the explanation length of PI-explanation varies from two to more than ten.}
\label{fig:path_distrib}    
\end{figure}
Table~\ref{table:depth_comparison} shows the comparison of the average explanation length of the model generated by our Cascading Decision Trees and PI-explanations.
Our Cascading Decision Trees algorithm shortens the explainable paths to users by \pathimprovement on average among ten datasets.
The average representation size of the Cascading Decision Trees model is \casdepth among ten datasets. 
We specifically focus on the explanations for positive classifications.
This means on average, only \casdepth features are necessary for the classification of a positive sample in Cascading Decision Trees.
On the contrary, for PI-explanations, \classicdepth features are necessary for a positive classification.
Figure~\ref{fig:path_distrib} shows the distribution of explanation lengths for all the examples in the Breast Cancer dataset.
Due to the page limit, we present the results of the other seven datasets in the supplemental material.
This succinctness of the explanation improves users' ability to diagnose the causes for the diseases and support scientific hypothesis.
We remark that PI-explanation could also be applied to Cascading Decision Tree.
Let $\text{PI}(f,x)$ be the PI-explanation of model $f$ on input $x$.
By the definition of PI-explanation, we know that $|\text{PI}(f,x)| \le |e^f(x;w)|$ always holds for all input $x$.
As shown in Table~\ref{table:depth_comparison}, $\mathbb{E}\left[|e^{f_\text{cascading}}(x;w)|\right] \le \mathbb{E}\left[|\text{PI}(f_\text{classic},x)|\right]$ holds.
Therefore we point out that
\begin{align*}
    \mathbb{E}\left[|\text{PI}(f_\text{cascading},x)|\right] &\le \mathbb{E}\left[|e^{f_\text{cascading}}(x;w)|\right] \\
    &\le \mathbb{E}\left[|\text{PI}(f_\text{classic},x)|\right]
\end{align*}

\begin{table*}[t]
\def\arraystretch{1.1}
\centering
\caption{Breakdown Evaluation of Cascading Decision Trees on CI Build Status Prediction.}
\label{table:casResultCI}
\resizebox{1.9\columnwidth}{!}{
\begin{footnotesize}
\begin{tabular}{|l|l|l|l|l|l|l|l|l|l|l|}
\hline
 Cascading & Explanation Depth  & Accuracy & Runtime (s) & TP & TN & FP & FN & Precision & Recall & F-1 Score \\ \hline
ON & 2.18  & 90.55\% & 249.43 &	135 &	1428 & 68  & 95  &	66.50\%	 & 58.70\% &	62.36\%   \\ \hline
OFF & 2.31 & 90.50\% & 277.95 &	135 &	1427 & 69  & 95  &	66.18\%	 & 58.70\% &	62.21\% \\ \hline

\end{tabular}
\end{footnotesize}
}
\end{table*}

\para{Accuracy.} 
Table~\ref{table:Result} shows the breakdown of the evaluation of Cascading Decision Trees.
In nine out of ten datasets, our Cascading Decision Trees algorithm surprisingly outperforms the PI-explanation in test accuracy.
This shows that long decision paths not only make models difficult to interpret, but also unavoidably lead to overfitting problems, which diminishes the overall test accuracy.
To lower the explanation path of the decision trees model, one standard technique for classic decision trees is to fix a maximum depth.
However, contrary to the Cascading Decision Trees, setting the maximum depth to three incurs a decrease in average prediction accuracy by around 2.0\%.
Even with this maximum depth and lower accuracy, in all ten datasets, the average explanation path is still longer than Cascading Decision Trees.
The comparison between Cascading Decision Trees with sequential covering manifests in the same vein.
In conclusion, compared with all three baselines, our Cascading Decision Trees delivers better model comprehensibility via shorter explanations while maintaining high prediction accuracy.

\para{Low False-positive Rate.}
The Cascading Decision Trees model has another key advantage - the low false-positive rate in prediction, which is crucial in medical and scientific domains.
For example, if the prediction result is positive (the presence of disease), the doctor needs to understand and explain the rationale behind the diagnosis of the disease and then report them to patients.
A high false-positive rate not only reduces physicians' confidence in adopting the prediction result but also leads to unnecessary and invasive follow-up tests on patients~\cite{HNRfalsepositive}.
Therefore, it is crucial to have a competitive accuracy prediction model with very low false-positive rate.
As shown in Table~\ref{table:Result}, the Cascading Decision Trees algorithm has the lowest false positive (FP) rate in nine of ten datasets compared to the PI-explanation.

\begin{figure}[t]
 \centering
      \resizebox{.80\linewidth}{!}{\begin{tikzpicture}
\begin{axis}[
    title={},
    xlabel={Missing data ratio (\%)},
    ylabel={Prediction failure rate (\%)},
    xmin=0, xmax=100,
    ymin=0, ymax=100,
    xtick={0,10,20,30,40,50,60,70,80,90,100},
    ytick={0,10,20,30,40,50,60,70,80,90,100},
    legend pos=south east,
    ymajorgrids=true,
    grid style=dashed,
]

\addplot+[only marks,
    error bars/.cd,
    y dir=both,
    y explicit,
    error bar style={line width=1pt},
    error mark options={
      rotate=90,
      red,
      mark size=4pt,
      line width=0.5pt
    }
    ]
    coordinates {
    (0,0) +- (0,0)
    (6.6,37) +- (0,0.7)
    (13.3,67) +- (0,8.2)
    (20,80) +- (0,6.7)
    (26.6,90) +- (0,3.3)
    (33.3,92.2) +- (0,1.1)
    (40,93.3) +- (0,1.1)
    (46.6,95.6) +- (0,1.0)
    (60,97.8) +- (0,1.0)
    (80,100) +- (0,0)
    (100,100) +- (0,0)
    };

\addplot+[only marks,
    error bars/.cd,
    y dir=both,
    y explicit,
    error bar style={line width=1pt},
    error mark options={
      rotate=90,
      red,
      mark size=4pt,
      line width=0.5pt
    }
    ]
    coordinates {
    (0,0) +- (0,0)
    (6.6,11) +- (0,0.4)
    (13.3,23) +- (0,2.7)
    (20,40) +- (0,2.9)
    (26.6,48.6) +- (0,5.7)
    (33.3,62) +- (0,7.8)
    (40,65.8) +- (0,2.7)
    (46.6,71.4) +- (0,5.7)
    (60,82.9) +- (0,2.7)
    (80,94.3) +- (0,2.7)
    (100,100) +- (0,0)
    };

\addplot[
    color=blue,
    mark=triangle,
    ]
    coordinates {
    (0,0)(6.6,37)(13.3,67)(20,80)(26.6,90)(33.3,92.2)(40,93.3)(46.6,95.6)(60,97.8)(80,100)(100,100)
    };

\addplot[
    color=red,
    mark=square,
    ]
    coordinates {
    (0,0) (6.6,11)(13.3,23)(20,40)(26.6,48.6)(33.3,62)(40,65.8)(46.6,71.4)(60,82.9)(80,94.3)(100,100)
    };
    \legend{,,PI-explanation,Cascading Decision Trees}

\end{axis}
\end{tikzpicture}}
      \caption{Robustness against missing data. Cascading Decision Trees vs. PI-explanation on the Breast Cancer dataset.}
      \label{fig:linegraph}
\end{figure}

\para{Robustness Against Missing Data.}
Our evaluation also demonstrates that Cascading Decision Tree has advantages in the robustness against missing data.
For the PI-explanation, if any feature along the selected decision path contains a missing value in the testing sample, the model stops immediately and fails to make an inference.
Therefore, the longer decision path is, the more likely it is for the inference to fail in the presence of missing data.

Based on the average \pathimprovement shorter decision paths than PI-explanation, we expect that Cascading Decision Tree will be robust against datasets with missing data.
To test the ability of missing data handling in practice, we randomly selected part of the testing data as missing.
Figure~\ref{fig:linegraph} confirms our expectation. illustrates that the Cascading Decision Tree has advantages in robustness against missing data in the Breast Cancer dataset.
Due to the page limit, we present the results of the other nine datasets in the supplemental material.

\para{Efficiency.} Although this paper does not focus on building a fast classifier, the efficiency of our Cascading Decision Trees algorithm turns out to be strong in practice. 
The training process finishes all ten real-world UCI datasets in seconds.
Taking the ionosphere dataset as an example, our Cascading Decision Trees learning process terminates in 1.1 seconds with accuracy of 88.7\%. 
This is comparable to accuracy of 87.0\% of the state-of-the-art optimal decision trees algorithm BioOCT when depth of the tree is three (cf.~\cite{VerwerZ19}). 
However, BioOCT takes around ten minutes to train.
Moreover, when the dataset size becomes larger, the Cascading Decision Trees scales better than the baselines.
The German credit dataset is the largest dataset in our evaluation. 
It takes the PI-explaination more than 50 mins to train.
Our Cascading Decision Trees training process not only terminates in 1.6 seconds, but also it comes with an advantage in average prediction accuracy around 5.0\%.

\para{Reproducibility.} Source code and the data is publicly available at:
\url{https://doi.org/10.5281/zenodo.7377693}.

\para{Generalizing Cascading Decision Tree to Real-world Applications.} We further apply the Cascading Decision Trees for predicting the build status in a novel real-world application, continuous integration (CI) build status prediction~\cite{10.1145/2568225.2568260}. 
In CI, users upload their code as one commit, and CI starts to build and test users' code under customized conditions such as the choice of operating system, library dependencies and other similar properties.
However, one major drawback of CI is that the CI build attempt can be extremely time consuming.
To address this issue, the existing work~\cite{santolucito2018statically} first collected the historical repository data stored in the CI environment. It then took advantage of the transparent structure of the decision tree model to predict the CI build status.

This CI build status predicting problem is an ideal testbed for our Cascading Decision Trees, because the main module behind the tool is a classic decision tree for binary classification. 
Moreover, when using decision tree in such a program analysis domain, having a very low false-positive rate is crucial for user acceptance of the tool~\cite{junker2012smt}. 
We contacted the authors, adopted the same dataset and ran the same study they used with only one exception - substituting Cascading Decision Trees for classic decision tree in predicting the CI build status.

Table~\ref{table:casResultCI} shows the breakdown evaluation of using Cascading Decision Trees on CI build status prediction.
The evaluation results show that the
Cascading Decision Trees provides a shorter explanation, with a competitive prediction accuracy of 90.55\%.
The Cascading Decision Trees model shortens the explanation length for failed builds by 5.6\%.
Moreover, the ratio of the average number of false positives reports to the average number of correct classifications is only 4.4\% (FP/TP+TN).
Our study shows that the use of Cascading Decision Trees provides developers with a more succinct but comprehensible set of rules that are responsible for positive classifications.

\section{Conclusions}

Learning decision trees on modern datasets generates large trees, which in turn produce decision paths of excessive depth, obscuring the explanation of classifications.
This paper intends to maximize model comprehensibility while maintaining prediction accuracy in binary classification.
The Cascading Decision Trees algorithm has been proposed to provide more succinct explanations in binary decision trees.
We evaluated our algorithm in real-world medical, scientific and program analysis datasets, where the explainability of the positive test result is of the utmost importance.
Our Cascading Decision Trees algorithm shortens the explanation depth by over \pathimprovement for positive classifications compared to the classic decision trees with PI-explanations.

\bibliography{os,travis,ai}

\newpage
\appendix
\onecolumn
\section*{Supplemental Material}
\label{sec:append}

\section{Algorithm.}
\label{subsec:alg}

\begin{algorithm}[h]
   \caption{Constuct Cascading Decision Trees Classifier}
   \label{algo:train}
\begin{algorithmic}
   \STATE {\bfseries Input:} $\trainingSet=\{(x,y)\}$: labelled training set 
   \STATE {\bfseries Input:} $\theta$: threshold for $mixed$ value
   \STATE {\bfseries Input:} $depth$: maximum depth of each decision tree
   \STATE {\bfseries Output:} $cascadingTree$: a list of cascading decision tree classifiers
  
   \STATE $done$ = False
   \STATE $cascadingTree$ = []
   \STATE $t = 0$, $\trainingSet_0 = \trainingSet$
   \WHILE {$done = False$}
   \STATE $clf$ = classicDecisionTree.fit($\trainingSet,depth$)
   \STATE $cascadingTree$.append($clf$)
   \STATE $done = \mathbbm{1}\left[\forall leaf \in clf, ~~mixed(leaf) < \theta\right]$
   \IF{$(done = False$)}
      \STATE $\trainingSet_{t+1} = \{(x,y)\in\trainingSet_t: clf.predict(x) = Neg\}$
   \ENDIF
   \STATE $t = t+1$
   \ENDWHILE
   \STATE {\bfseries return} $cascadingTree$
   
\end{algorithmic}
\end{algorithm}

\begin{algorithm}[h]
      \caption{Cascading Decision Trees Inference}
      \label{algo:test}
      \begin{algorithmic}
            \STATE {\bfseries Input:} $x\in\mathcal{X}$
            \STATE {\bfseries Input:} cascadingTreeClassifier: $cascadingTree$
            \STATE {\bfseries Output:} prediction: A Boolean, Postive or Negative.
            \STATE {\bfseries Output:} explanation paths: A conjunction of boolean statements to explain the postive classifications. 
            
            \FOR {$clf$ {\bfseries in} $cascadingTree$ }
            \STATE decisionPath = clf.decisionPath($sample,\theta$)
            \IF { $sample$.missValue {\bfseries in} decisionPath}
            \STATE {\bfseries continue}
            \ELSE
            \IF{ $clf.predict($sample$,\theta) = Positive$}
            \STATE {\bfseries return} (Postive, $clf$.path(sample))
            \ENDIF
            \ENDIF
            \ENDFOR
            \STATE {\bfseries return} (Negative)
            
      \end{algorithmic}
\end{algorithm}

\section{Datasets Breakdown} 

As detailed below, analysis of each of the following ten UCI datasets clearly benefit from more succinct explanations for positive predictions. They are:

\subsection{Breast Cancer}
\label{subsec:breast}
Predicting whether the reoccurrence of breast cancer happens. ``Positive'' samples mean the reoccurrence is observed, while ``Negative'' samples means there is no sign of the reoccurrence of cancer.

The distribution of explanation lengths for all the examples in the Breast Cancer dataset is presented in Figure~\ref{fig:path_distrib_breast}. 
The robustness against missing data is demonstrated in Figure~\ref{fig:linegraph_breast}.


    \pgfplotsset{
        compat=1.9,
        %
        blank pyramid axis style/.style={
            width=0.3*\textwidth,
            height=0.3*\textheight,
            scale only axis,
            xmin=0,
            xmax=100,
            ymin=-0.5,
            ymax=9,
            y dir=reverse,
            enlarge y limits={value=0.075,upper},
            xbar,
            axis x line=left,
            xtick align=outside,
            bar width=1,
            allow reversal of rel axis cs=false,
        },
        pyramid axis style/.style={
            blank pyramid axis style,
            %
            xticklabel={%
                \pgfmathprintnumber\tick\%%
            },
            ytick=\empty,
            axis line style={-},
            %
            nodes near coords={%
                \pgfmathprintnumber\pgfplotspointmeta\%%
            },
            every node near coord/.append style={
                font=\small,
                color=black,
                /pgf/number format/fixed,
            },
        },
    }
\begin{figure}[h]
\centering
    \begin{tikzpicture}[scale=0.63]
        \pgfplotstableread[
            col sep=comma,
            header=true,
        ]{
            distribution,cascading,classic
            +10,0,20
            9--10,0,2
            8--9,0,10
            7--8,0,7
            6--7,0,16
            5--6,0,12
            4--5,0,6
            3--4,0,11
            2--3,0,6
            1--2,35,0
        }\loadedtable
        \pgfplotstablecreatecol[
            expr accum={
                round(\pgfmathaccuma) + \thisrow{cascading} + \thisrow{classic}
            }{0}
        ]{sum}{\loadedtable}
        \tikzset{
            fpu=true,
        }
            \pgfplotstablegetrowsof{\loadedtable}
                \pgfmathsetmacro{\LastRow}{\pgfplotsretval-1}
            \pgfplotstablegetelem{9}{sum}\of{\loadedtable}
                \pgfmathsetmacro{\Sum}{\pgfplotsretval}
        \tikzset{
            fpu=false,
        }
        \begin{axis}[
            pyramid axis style,
            %
            axis y line*=left,
            ytick=\empty,
            name=popaxis,
        ]
            \addplot [cyan,fill=cyan!70] table [
                y expr =\coordindex,x expr={\thisrow{classic}/90*100},
            ] \loadedtable;

            \node [anchor=south] at (rel axis cs:0.65,0.6)
                {\textcolor{cyan}{PI-explanation}};
        \end{axis}

        \begin{axis}[
            pyramid axis style,
            %
            at={(popaxis.west)},
            anchor=east,
            xshift=-12.5mm,
            %
            x dir=reverse,
            every node near coord/.append style={
                anchor=east,
            },
            axis y line*=right,
        ]
            \addplot [teal,fill=teal!65] table [
                y expr =\coordindex, x expr={\thisrow{cascading}/\thisrow{cascading}*100},
            ] \loadedtable;

            \node [anchor=south] at (rel axis cs:0.6,0.6)
                {\textcolor{teal}{Cascading Decision Trees}};
        \end{axis}

        \begin{axis}[
            blank pyramid axis style,
            %
            at={(popaxis.west)},
            anchor=east,
            xshift=-12.5mm,
            %
            x dir=reverse,
            axis y line*=right,
            xtick=\empty,
            ytick=data,
            yticklabels from table={\loadedtable}{distribution},
            y tick label style={
                align=center,
                inner sep=0pt,
                text width=12.5mm,
            },
            major tick length=0pt,
            axis line style={
                -,
                draw=none,
            },
        ]
            \addplot [draw=none,fill=none] table [
                y expr =\coordindex, x expr={0},
            ] \loadedtable;

        \end{axis}
    \end{tikzpicture}
\caption{A distribution of explanation lengths for all the examples in the Breast Cancer dataset.}
\label{fig:path_distrib_breast}    
\end{figure}
\begin{figure}[h!]
 \centering
      \resizebox{.4\linewidth}{!}{\begin{tikzpicture}
\begin{axis}[
    title={},
    xlabel={Missing data ratio (\%)},
    ylabel={Prediction failure rate (\%)},
    xmin=0, xmax=100,
    ymin=0, ymax=100,
    xtick={0,10,20,30,40,50,60,70,80,90,100},
    ytick={0,10,20,30,40,50,60,70,80,90,100},
    legend pos=south east,
    ymajorgrids=true,
    grid style=dashed,
]

\addplot+[only marks,
    error bars/.cd,
    y dir=both,
    y explicit,
    error bar style={line width=1pt},
    error mark options={
      rotate=90,
      red,
      mark size=4pt,
      line width=0.5pt
    }
    ]
    coordinates {
    (0,0) +- (0,0)
    (6.6,37) +- (0,0.7)
    (13.3,67) +- (0,8.2)
    (20,80) +- (0,6.7)
    (26.6,90) +- (0,3.3)
    (33.3,92.2) +- (0,1.1)
    (40,93.3) +- (0,1.1)
    (46.6,95.6) +- (0,1.0)
    (60,97.8) +- (0,1.0)
    (80,100) +- (0,0)
    (100,100) +- (0,0)
    };

\addplot+[only marks,
    error bars/.cd,
    y dir=both,
    y explicit,
    error bar style={line width=1pt},
    error mark options={
      rotate=90,
      red,
      mark size=4pt,
      line width=0.5pt
    }
    ]
    coordinates {
    (0,0) +- (0,0)
    (6.6,11) +- (0,0.4)
    (13.3,23) +- (0,2.7)
    (20,40) +- (0,2.9)
    (26.6,48.6) +- (0,5.7)
    (33.3,62) +- (0,7.8)
    (40,65.8) +- (0,2.7)
    (46.6,71.4) +- (0,5.7)
    (60,82.9) +- (0,2.7)
    (80,94.3) +- (0,2.7)
    (100,100) +- (0,0)
    };

\addplot[
    color=blue,
    mark=triangle,
    ]
    coordinates {
    (0,0)(6.6,37)(13.3,67)(20,80)(26.6,90)(33.3,92.2)(40,93.3)(46.6,95.6)(60,97.8)(80,100)(100,100)
    };

\addplot[
    color=red,
    mark=square,
    ]
    coordinates {
    (0,0) (6.6,11)(13.3,23)(20,40)(26.6,48.6)(33.3,62)(40,65.8)(46.6,71.4)(60,82.9)(80,94.3)(100,100)
    };
    \legend{,,PI-explanation,Cascading Decision Trees}

\end{axis}
\end{tikzpicture}}
      \caption{Robustness against missing data. Cascading Decision Tree vs. PI-explanation in the Breast Cancer dataset.}
      \label{fig:linegraph_breast}
\end{figure}

\newpage
\subsection{Heart Disease}
\label{subsec:heart}
Predicting absence or presence of heart disease. ``Positive'' samples means the presence of heart disease, while ``Negative'' samples represent the absence.

The distribution of explanation lengths for all the examples in the Heart Disease dataset is presented in Figure~\ref{fig:path_distrib_heart}. 
The robustness against missing data is demonstrated in Figure~\ref{fig:linegraph_heart}.


    \pgfplotsset{
        compat=1.9,
        %
        blank pyramid axis style/.style={
            width=0.3*\textwidth,
            height=0.3*\textheight,
            scale only axis,
            xmin=0,
            xmax=100,
            ymin=-0.5,
            ymax=9,
            y dir=reverse,
            enlarge y limits={value=0.075,upper},
            xbar,
            axis x line=left,
            xtick align=outside,
            bar width=1,
            allow reversal of rel axis cs=false,
        },
        pyramid axis style/.style={
            blank pyramid axis style,
            %
            xticklabel={%
                \pgfmathprintnumber\tick\%%
            },
            ytick=\empty,
            axis line style={-},
            %
            nodes near coords={%
                \pgfmathprintnumber\pgfplotspointmeta\%%
            },
            every node near coord/.append style={
                font=\small,
                color=black,
                /pgf/number format/fixed,
            },
        },
    }
\begin{figure}[h!]
     \centering
    \begin{tikzpicture}[scale=0.63]
        \pgfplotstableread[
            col sep=comma,
            header=true,
        ]{
            distribution,cascading,classic
            +10,0,0
            9--10,0,0
            8--9,0,1
            7--8,0,2
            6--7,0,7
            5--6,0,7
            4--5,0,20
            3--4,0,41
            2--3,0,33
            1--2,74,0
        }\loadedtable
        \pgfplotstablecreatecol[
            expr accum={
                round(\pgfmathaccuma) + \thisrow{cascading} + \thisrow{classic}
            }{0}
        ]{sum}{\loadedtable}
        \tikzset{
            fpu=true,
        }
            \pgfplotstablegetrowsof{\loadedtable}
                \pgfmathsetmacro{\LastRow}{\pgfplotsretval-1}
            \pgfplotstablegetelem{9}{sum}\of{\loadedtable}
                \pgfmathsetmacro{\Sum}{\pgfplotsretval}
        \tikzset{
            fpu=false,
        }
        \begin{axis}[
            pyramid axis style,
            %
            axis y line*=left,
            ytick=\empty,
            name=popaxis,
        ]
            \addplot [cyan,fill=cyan!70] table [
                y expr =\coordindex,x expr={\thisrow{classic}/111*100},
            ] \loadedtable;

            \node [anchor=south] at (rel axis cs:0.65,0.6)
                {\textcolor{cyan}{PI-explanation}};
        \end{axis}

        \begin{axis}[
            pyramid axis style,
            %
            at={(popaxis.west)},
            anchor=east,
            xshift=-12.5mm,
            %
            x dir=reverse,
            every node near coord/.append style={
                anchor=east,
            },
            axis y line*=right,
        ]
            \addplot [teal,fill=teal!65] table [
                y expr =\coordindex, x expr={\thisrow{cascading}/74*100},
            ] \loadedtable;

            \node [anchor=south] at (rel axis cs:0.6,0.6)
                {\textcolor{teal}{Cascading Decision Trees}};
        \end{axis}

        \begin{axis}[
            blank pyramid axis style,
            %
            at={(popaxis.west)},
            anchor=east,
            xshift=-12.5mm,
            %
            x dir=reverse,
            axis y line*=right,
            xtick=\empty,
            ytick=data,
            yticklabels from table={\loadedtable}{distribution},
            y tick label style={
                align=center,
                inner sep=0pt,
                text width=12.5mm,
            },
            major tick length=0pt,
            axis line style={
                -,
                draw=none,
            },
        ]
            \addplot [draw=none,fill=none] table [
                y expr =\coordindex, x expr={0},
            ] \loadedtable;

        \end{axis}
    \end{tikzpicture}
\caption{A distribution of explanation lengths for all the examples in the Heart Disease dataset.}
\label{fig:path_distrib_heart}    
\end{figure}
\begin{figure}[h!]
 \centering
      \resizebox{.4\linewidth}{!}{\begin{tikzpicture}
\begin{axis}[
    title={},
    xlabel={Missing data ratio (\%)},
    ylabel={Prediction failure rate (\%)},
    xmin=0, xmax=100,
    ymin=0, ymax=100,
    xtick={0,10,20,30,40,50,60,70,80,90,100},
    ytick={0,10,20,30,40,50,60,70,80,90,100},
    legend pos=south east,
    ymajorgrids=true,
    grid style=dashed,
]

\addplot+[only marks,
    error bars/.cd,
    y dir=both,
    y explicit,
    error bar style={line width=1pt},
    error mark options={
      rotate=90,
      red,
      mark size=4pt,
      line width=0.5pt
    }
    ]
    coordinates {
    (0,0) +- (0,0)
    (7.7,35.1) +- (0,5.4)
    (15.4,55.8) +- (0,5.4)
    (23.1,64.8) +- (0,2.7)
    (30.77,79.2) +- (0,6.3)
    (38.5,83.7) +- (0,6.3)
    (46.2,94.5) +- (0,3.6)
    (61.5,97.2) +- (0,1.8)
    (76.9,100) +- (0,0)
    (100,100) +- (0,0)
    };

\addplot+[only marks,
    error bars/.cd,
    y dir=both,
    y explicit,
    error bar style={line width=1pt},
    error mark options={
      rotate=90,
      red,
      mark size=4pt,
      line width=0.5pt
    }
    ]
    coordinates {
    (0,0) +- (0,0)
    (7.7,10.8) +- (0,5.4)
    (15.4,27) +- (0,9.4)
    (23.1,35.1) +- (0,12.1)
    (30.77,54) +- (0,2.7)
    (38.5,63.5) +- (0,4.0)
    (46.2,70.2) +- (0,4.1)
    (61.5,78.3) +- (0,6.8)
    (76.9,95.9) +- (0,1.3)
    (100,100) +- (0,0)
    };

\addplot[
    color=blue,
    mark=triangle,
    ]
    coordinates {
    (0,0)(7.7,35.1)(15.4,55.8)(23.1,64.8)(30.77,79.2)(38.5,83.7)(46.2,94.5)(61.5,99)(76.9,100)(100,100)
    
    };

\addplot[
    color=red,
    mark=square,
    ]
    coordinates {
    (0,0)(7.7,10.8)(15.4,27)(23.1,35.1)(30.77,54)(38.5,63.5)(46.2,70.2)(61.5,78.3)(76.9,97.2)(100,100)
    };
    \legend{,,PI-explanation,Cascading Decision Trees}
    
\end{axis}
\end{tikzpicture}}
      \caption{Robustness against missing data. Cascading Decision Tree vs. PI-explanation in the Heart Disease dataset.}
      \label{fig:linegraph_heart}
\end{figure}

\newpage
\subsection{Wisc Breast Cancer}
\label{wisc}
Classification of breast cancer. ``Positive'' samples mean the tumor turns out to be malignant, while ``Negative'' samples are benign.

The distribution of explanation lengths for all the examples in the Wisc Breast Cancer dataset is presented in Figure~\ref{fig:path_distrib_wisc}. 
The robustness against missing data is demonstrated in Figure~\ref{fig:linegraph_wisc}.


    \pgfplotsset{
        compat=1.9,
        %
        blank pyramid axis style/.style={
            width=0.3*\textwidth,
            height=0.3*\textheight,
            scale only axis,
            xmin=0,
            xmax=100,
            ymin=-0.5,
            ymax=9,
            y dir=reverse,
            enlarge y limits={value=0.075,upper},
            xbar,
            axis x line=left,
            xtick align=outside,
            bar width=1,
            allow reversal of rel axis cs=false,
        },
        pyramid axis style/.style={
            blank pyramid axis style,
            %
            xticklabel={%
                \pgfmathprintnumber\tick\%%
            },
            ytick=\empty,
            axis line style={-},
            %
            nodes near coords={%
                \pgfmathprintnumber\pgfplotspointmeta\%%
            },
            every node near coord/.append style={
                font=\small,
                color=black,
                /pgf/number format/fixed,
            },
        },
    }
\begin{figure}[h!]
    \centering
    \begin{tikzpicture}[scale=0.63]
        \pgfplotstableread[
            col sep=comma,
            header=true,
        ]{
            distribution,cascading,classic
            +10,0,0
            9--10,0,0
            8--9,0,0
            7--8,0,0
            6--7,0,1
            5--6,0,1
            4--5,0,7
            3--4,0,34
            2--3,0,38
            1--2,191,130
        }\loadedtable
        \pgfplotstablecreatecol[
            expr accum={
                round(\pgfmathaccuma) + \thisrow{cascading} + \thisrow{classic}
            }{0}
        ]{sum}{\loadedtable}
        \tikzset{
            fpu=true,
        }
            \pgfplotstablegetrowsof{\loadedtable}
                \pgfmathsetmacro{\LastRow}{\pgfplotsretval-1}
            \pgfplotstablegetelem{9}{sum}\of{\loadedtable}
                \pgfmathsetmacro{\Sum}{\pgfplotsretval}
        \tikzset{
            fpu=false,
        }
        \begin{axis}[
            pyramid axis style,
            %
            axis y line*=left,
            ytick=\empty,
            name=popaxis,
        ]
            \addplot [cyan,fill=cyan!70] table [
                y expr =\coordindex,x expr={\thisrow{classic}/191*100},
            ] \loadedtable;

            \node [anchor=south] at (rel axis cs:0.65,0.6)
                {\textcolor{cyan}{PI-explanation}};
        \end{axis}

        \begin{axis}[
            pyramid axis style,
            %
            at={(popaxis.west)},
            anchor=east,
            xshift=-12.5mm,
            %
            x dir=reverse,
            every node near coord/.append style={
                anchor=east,
            },
            axis y line*=right,
        ]
            \addplot [teal,fill=teal!65] table [
                y expr =\coordindex, x expr={\thisrow{cascading}/191*100},
            ] \loadedtable;

            \node [anchor=south] at (rel axis cs:0.6,0.6)
                {\textcolor{teal}{Cascading Decision Trees}};
        \end{axis}

        \begin{axis}[
            blank pyramid axis style,
            %
            at={(popaxis.west)},
            anchor=east,
            xshift=-12.5mm,
            %
            x dir=reverse,
            axis y line*=right,
            xtick=\empty,
            ytick=data,
            yticklabels from table={\loadedtable}{distribution},
            y tick label style={
                align=center,
                inner sep=0pt,
                text width=12.5mm,
            },
            major tick length=0pt,
            axis line style={
                -,
                draw=none,
            },
        ]
            \addplot [draw=none,fill=none] table [
                y expr =\coordindex, x expr={0},
            ] \loadedtable;

        \end{axis}
    \end{tikzpicture}
\caption{A distribution of explanation lengths for all the examples in the Wisc Breast Cancer dataset.}
\label{fig:path_distrib_wisc}    
\end{figure}
\begin{figure}[h!]
 \centering
      \resizebox{.4\linewidth}{!}{\begin{tikzpicture}
\begin{axis}[
    title={},
    xlabel={Missing data ratio (\%)},
    ylabel={Prediction failure rate (\%)},
    xmin=0, xmax=100,
    ymin=0, ymax=100,
    xtick={0,10,20,30,40,50,60,70,80,90,100},
    ytick={0,10,20,30,40,50,60,70,80,90,100},
    legend pos=south east,
    ymajorgrids=true,
    grid style=dashed,
]

\addplot+[only marks,
    error bars/.cd,
    y dir=both,
    y explicit,
    error bar style={line width=1pt},
    error mark options={
      rotate=90,
      red,
      mark size=4pt,
      line width=0.5pt
    }
    ]
    coordinates {
    (0,0) +- (0,0)
    (3.4,7.1) +- (0,0.5)
    (10.3,21.2) +- (0,3.4)
    (17.2,36) +- (0,3.8)
    (31.0,59.2) +- (0,2.7)
    (43.1,76.8) +- (0,1.3)
    (55.2,84.4) +- (0,0.045)
    (75.9,94.8) +- (0,1.0)
    (86.2,98.6) +- (0,0.4)
    (100,100) +- (0,0)
    };

\addplot+[only marks,
    error bars/.cd,
    y dir=both,
    y explicit,
    error bar style={line width=1pt},
    error mark options={
      rotate=90,
      red,
      mark size=4pt,
      line width=0.5pt
    }
    ]
    coordinates {
    (0,0) +- (0,0)
    (3.4,3.6) +- (0,1.6)
    (10.3,17.6) +- (0,4.5)
    (17.2,34.7) +- (0,4.2)
    (31.0,51.8) +- (0,2.0)
    (43.1,70.8) +- (0,3.4)
    (55.2,81.3) +- (0,1.6)
    (75.9,93.3) +- (0,1.6)
    (86.2,97.9) +- (0,1.1)
    (100,100) +- (0,0)
    };

\addplot[
    color=blue,
    mark=triangle,
    ]
    coordinates {
    (0,0)(3.4,7.1)(10.3,21.2)(17.2,36)(31.0,59.2)(43.1,76.8)(55.2,84.4)(75.9,94.8)(86.2,98.6)(100,100)
    
    };

\addplot[
    color=red,
    mark=square,
    ]
    coordinates {
    (0,0)(3.4,5.2)(10.3,17.6)(17.2,34.7)(31.0,51.8)(43.1,74.1)(55.2,81.3)(75.9,93.3)(86.2,99)(100,100)
    };
    \legend{,,PI-explanation,Cascading Decision Trees}
    
\end{axis}
\end{tikzpicture}}
      \caption{Robustness against missing data. Cascading Decision Tree vs. PI-explanation in the Wisc Breast Cancer dataset.}
      \label{fig:linegraph_wisc}
\end{figure}

\newpage
\subsection{Cervical}
\label{subsec:cervical}
Predicting absence or presence of cervical cancer. ``Positive'' samples means the presence of heart disease, while ``Negative'' samples represent the absence.

The distribution of explanation lengths for all the examples in the Cervical dataset is presented in Figure~\ref{fig:path_distrib_cervical}. 
The robustness against missing data is demonstrated in Figure~\ref{fig:linegraph_cervical}.


    \pgfplotsset{
        compat=1.9,
        %
        blank pyramid axis style/.style={
            width=0.3*\textwidth,
            height=0.3*\textheight,
            scale only axis,
            xmin=0,
            xmax=100,
            ymin=-0.5,
            ymax=9,
            y dir=reverse,
            enlarge y limits={value=0.075,upper},
            xbar,
            axis x line=left,
            xtick align=outside,
            bar width=1,
            allow reversal of rel axis cs=false,
        },
        pyramid axis style/.style={
            blank pyramid axis style,
            %
            xticklabel={%
                \pgfmathprintnumber\tick\%%
            },
            ytick=\empty,
            axis line style={-},
            %
            nodes near coords={%
                \pgfmathprintnumber\pgfplotspointmeta\%%
            },
            every node near coord/.append style={
                font=\small,
                color=black,
                /pgf/number format/fixed,
            },
        },
    }
\begin{figure}[h!]
\centering
    \begin{tikzpicture}[scale=0.63]
        \pgfplotstableread[
            col sep=comma,
            header=true,
        ]{
            distribution,cascading,classic
            +10,0,6
            9--10,0,1
            8--9,0,3
            7--8,0,1
            6--7,0,14
            5--6,0,5
            4--5,0,7
            3--4,0,0
            2--3,0,2
            1--2,61,0
        }\loadedtable
        \pgfplotstablecreatecol[
            expr accum={
                round(\pgfmathaccuma) + \thisrow{cascading} + \thisrow{classic}
            }{0}
        ]{sum}{\loadedtable}
        \tikzset{
            fpu=true,
        }
            \pgfplotstablegetrowsof{\loadedtable}
                \pgfmathsetmacro{\LastRow}{\pgfplotsretval-1}
            \pgfplotstablegetelem{9}{sum}\of{\loadedtable}
                \pgfmathsetmacro{\Sum}{\pgfplotsretval}
        \tikzset{
            fpu=false,
        }
        \begin{axis}[
            pyramid axis style,
            %
            axis y line*=left,
            ytick=\empty,
            name=popaxis,
        ]
            \addplot [cyan,fill=cyan!70] table [
                y expr =\coordindex,x expr={\thisrow{classic}/39*100},
            ] \loadedtable;

            \node [anchor=south] at (rel axis cs:0.65,0.6)
                {\textcolor{cyan}{PI-explanation}};
        \end{axis}

        \begin{axis}[
            pyramid axis style,
            %
            at={(popaxis.west)},
            anchor=east,
            xshift=-12.5mm,
            %
            x dir=reverse,
            every node near coord/.append style={
                anchor=east,
            },
            axis y line*=right,
        ]
            \addplot [teal,fill=teal!65] table [
                y expr =\coordindex, x expr={\thisrow{cascading}/61*100},
            ] \loadedtable;

            \node [anchor=south] at (rel axis cs:0.6,0.6)
                {\textcolor{teal}{Cascading Decision Trees}};
        \end{axis}

        \begin{axis}[
            blank pyramid axis style,
            %
            at={(popaxis.west)},
            anchor=east,
            xshift=-12.5mm,
            %
            x dir=reverse,
            axis y line*=right,
            xtick=\empty,
            ytick=data,
            yticklabels from table={\loadedtable}{distribution},
            y tick label style={
                align=center,
                inner sep=0pt,
                text width=12.5mm,
            },
            major tick length=0pt,
            axis line style={
                -,
                draw=none,
            },
        ]
            \addplot [draw=none,fill=none] table [
                y expr =\coordindex, x expr={0},
            ] \loadedtable;

        \end{axis}
    \end{tikzpicture}
\caption{A distribution of explanation lengths for all the examples in the Cervical dataset.}
\label{fig:path_distrib_cervical}    
\end{figure}

\begin{figure}[h!]
 \centering
      \resizebox{.4\linewidth}{!}{\begin{tikzpicture}
\begin{axis}[
    title={},
    xlabel={Missing data ratio (\%)},
    ylabel={Prediction failure rate (\%)},
    xmin=0, xmax=100,
    ymin=0, ymax=100,
    xtick={0,10,20,30,40,50,60,70,80,90,100},
    ytick={0,10,20,30,40,50,60,70,80,90,100},
    legend pos=south east,
    ymajorgrids=true,
    grid style=dashed,
]

\addplot+[only marks,
    error bars/.cd,
    y dir=both,
    y explicit,
    error bar style={line width=1pt},
    error mark options={
      rotate=90,
      red,
      mark size=4pt,
      line width=0.5pt
    }
    ]
    coordinates {
    (0,0) +- (0,0)
    (3.0,23.0) +- (0,1.3)
    (6.0,25.7) +- (0,5.1)
    (9.0,51.3) +- (0,2.6)
    (15.5,73) +- (0,9.0)
    (30.3,94.9) +- (0,2.6)
    (51.5,97.4) +- (0,2.6)
    (100,100) +- (0,0)
    };

\addplot+[only marks,
    error bars/.cd,
    y dir=both,
    y explicit,
    error bar style={line width=1pt},
    error mark options={
      rotate=90,
      red,
      mark size=4pt,
      line width=0.5pt
    }
    ]
    coordinates {
    (0,0) +- (0,0)
    (3.3,4.9) +- (0,0.8)
    (6.0,14.8) +- (0,1.6)
    (9.0,18.0) +- (0,1.6)
    (15.5,23.8) +- (0,5.7)
    (30.3,49.2) +- (0,3.3)
    (51.5,74.6) +- (0,4.1)
    (100,100) +- (0,0)
    };

\addplot[
    color=blue,
    mark=triangle,
    ]
    coordinates {
    (0,0)(3.0,23.0)(6.0,25.7)(9.0,51.3)(15.5,73)(30.3,94.9)(51.5,97.4)(100,100)

    };

\addplot[
    color=red,
    mark=square,
    ]
    coordinates {
    (0,0)(3.3,4.9)(6.0,14.8)(9.0,18.0)(15.5,23.8)(30.3,49.2)(51.5,74.6)(100,100)
    };
    \legend{,,PI-explanation,Cascading Decision Trees}
    
\end{axis}
\end{tikzpicture}}
      \caption{Robustness against missing data. Cascading Decision Tree vs. PI-explanation in the Cervical Cancer dataset.}
      \label{fig:linegraph_cervical}
\end{figure}

\newpage
\subsection{Credit Card}
\label{subsec:credit}
Determining people who will be granted credit. ``Positive'' samples mean the credit is approved, while ``Negative'' samples mean the credit is disapproved.

The distribution of explanation lengths for all the examples in the Credit Card dataset is presented in Figure~\ref{fig:path_distrib_credit}. 
The robustness against missing data is demonstrated in Figure~\ref{fig:linegraph_credit}.


    \pgfplotsset{
        compat=1.9,
        %
        blank pyramid axis style/.style={
            width=0.3*\textwidth,
            height=0.3*\textheight,
            scale only axis,
            xmin=0,
            xmax=100,
            ymin=-0.5,
            ymax=9,
            y dir=reverse,
            enlarge y limits={value=0.075,upper},
            xbar,
            axis x line=left,
            xtick align=outside,
            bar width=1,
            allow reversal of rel axis cs=false,
        },
        pyramid axis style/.style={
            blank pyramid axis style,
            %
            xticklabel={%
                \pgfmathprintnumber\tick\%%
            },
            ytick=\empty,
            axis line style={-},
            %
            nodes near coords={%
                \pgfmathprintnumber\pgfplotspointmeta\%%
            },
            every node near coord/.append style={
                font=\small,
                color=black,
                /pgf/number format/fixed,
            },
        },
    }
\begin{figure}[h!]
\centering
    \begin{tikzpicture}[scale=0.63]
        \pgfplotstableread[
            col sep=comma,
            header=true,
        ]{
            distribution,cascading,classic
            +10,0,24
            9--10,0,7
            8--9,0,10
            7--8,0,35
            6--7,0,49
            5--6,0,94
            4--5,0,129
            3--4,0,14
            2--3,0,0
            1--2,306,0
        }\loadedtable
        \pgfplotstablecreatecol[
            expr accum={
                round(\pgfmathaccuma) + \thisrow{cascading} + \thisrow{classic}
            }{0}
        ]{sum}{\loadedtable}
        \tikzset{
            fpu=true,
        }
            \pgfplotstablegetrowsof{\loadedtable}
                \pgfmathsetmacro{\LastRow}{\pgfplotsretval-1}
            \pgfplotstablegetelem{9}{sum}\of{\loadedtable}
                \pgfmathsetmacro{\Sum}{\pgfplotsretval}
        \tikzset{
            fpu=false,
        }
        \begin{axis}[
            pyramid axis style,
            %
            axis y line*=left,
            ytick=\empty,
            name=popaxis,
        ]
            \addplot [cyan,fill=cyan!70] table [
                y expr =\coordindex,x expr={\thisrow{classic}/362*100},
            ] \loadedtable;

            \node [anchor=south] at (rel axis cs:0.65,0.6)
                {\textcolor{cyan}{PI-explanation}};
        \end{axis}

        \begin{axis}[
            pyramid axis style,
            %
            at={(popaxis.west)},
            anchor=east,
            xshift=-12.5mm,
            %
            x dir=reverse,
            every node near coord/.append style={
                anchor=east,
            },
            axis y line*=right,
        ]
            \addplot [teal,fill=teal!65] table [
                y expr =\coordindex, x expr={\thisrow{cascading}/306*100},
            ] \loadedtable;

            \node [anchor=south] at (rel axis cs:0.6,0.6)
                {\textcolor{teal}{Cascading Decision Trees}};
        \end{axis}

        \begin{axis}[
            blank pyramid axis style,
            %
            at={(popaxis.west)},
            anchor=east,
            xshift=-12.5mm,
            %
            x dir=reverse,
            axis y line*=right,
            xtick=\empty,
            ytick=data,
            yticklabels from table={\loadedtable}{distribution},
            y tick label style={
                align=center,
                inner sep=0pt,
                text width=12.5mm,
            },
            major tick length=0pt,
            axis line style={
                -,
                draw=none,
            },
        ]
            \addplot [draw=none,fill=none] table [
                y expr =\coordindex, x expr={0},
            ] \loadedtable;

        \end{axis}
    \end{tikzpicture}
\caption{A distribution of explanation lengths for all the examples in the Credit Card dataset.}
\label{fig:path_distrib_credit}    
\end{figure}
\begin{figure}[h!]
 \centering
      \resizebox{.4\linewidth}{!}{\begin{tikzpicture}
\begin{axis}[
    title={},
    xlabel={Missing data ratio (\%)},
    ylabel={Prediction failure rate (\%)},
    xmin=0, xmax=100,
    ymin=0, ymax=100,
    xtick={0,10,20,30,40,50,60,70,80,90,100},
    ytick={0,10,20,30,40,50,60,70,80,90,100},
    legend pos=south east,
    ymajorgrids=true,
    grid style=dashed,
]

\addplot+[only marks,
    error bars/.cd,
    y dir=both,
    y explicit,
    error bar style={line width=1pt},
    error mark options={
      rotate=90,
      red,
      mark size=4pt,
      line width=0.5pt
    }
    ]
    coordinates {
    (0,0) +- (0,0)
    (2.4,12.2) +- (0,3.3)
    (9.5,44.5) +- (0,3.3)
    (19,72.5) +- (0,5.7)
    (28.6,82.6) +- (0,2.4)
    (42.9,93.1) +- (0,1.4)
    (59.5,98.8) +- (0,0.4)
    (78.6,99.7) +- (0,0.3)
    (100,100) +- (0,0)
    };

\addplot+[only marks,
    error bars/.cd,
    y dir=both,
    y explicit,
    error bar style={line width=1pt},
    error mark options={
      rotate=90,
      red,
      mark size=4pt,
      line width=0.5pt
    }
    ]
    coordinates {
    (0,0) +- (0,0)
    (2.4,5.1) +- (0,1.5)
    (9.5,17.3) +- (0,2.3)
    (19,33.7) +- (0,1.6)
    (28.6,49.4) +- (0,2.0)
    (42.9,65) +- (0,4.0)
    (59.5,83) +- (0,1.3)
    (78.6,95.9) +- (0,0.5)
    (100,100) +- (0,0)
    };

\addplot[
    color=blue,
    mark=triangle,
    ]
    coordinates {
    (0,0)(2.4,12.2)(9.5,44.5)(19,72.5)(28.6,82.6)(42.9,93.1)(59.5,98.8)(78.6,99.7)(100,100)
    
    };

\addplot[
    color=red,
    mark=square,
    ]
    coordinates {
    (0,0)(2.4,5.1)(9.5,17.3)(19,33.7)(28.6,49.4)(42.9,65)(59.5,83)(78.6,95.9)(100,100)
    };
    \legend{,,PI-explanation,Cascading Decision Trees}
    
\end{axis}
\end{tikzpicture}}
      \caption{Robustness against missing data. Cascading Decision Tree vs. PI-explanation in the Credit Card dataset.}
      \label{fig:linegraph_credit}
\end{figure}

\newpage
\subsection{German Credit}
\label{subsec:german}
Infering each person's credit condition. ``Positive'' samples mean the person's credit is good, while ``Negative'' samples means the opposite.

The distribution of explanation lengths for all the examples in the German Credit dataset is presented in Figure~\ref{fig:path_distrib_german}. 
The robustness against missing data is demonstrated in Figure~\ref{fig:linegraph_german}.


    \pgfplotsset{
        compat=1.9,
        %
        blank pyramid axis style/.style={
            width=0.3*\textwidth,
            height=0.3*\textheight,
            scale only axis,
            xmin=0,
            xmax=100,
            ymin=-0.5,
            ymax=9,
            y dir=reverse,
            enlarge y limits={value=0.075,upper},
            xbar,
            axis x line=left,
            xtick align=outside,
            bar width=1,
            allow reversal of rel axis cs=false,
        },
        pyramid axis style/.style={
            blank pyramid axis style,
            %
            xticklabel={%
                \pgfmathprintnumber\tick\%%
            },
            ytick=\empty,
            axis line style={-},
            %
            nodes near coords={%
                \pgfmathprintnumber\pgfplotspointmeta\%%
            },
            every node near coord/.append style={
                font=\small,
                color=black,
                /pgf/number format/fixed,
            },
        },
    }
\begin{figure}[h!]
\centering
    \begin{tikzpicture}[scale=0.63]
        \pgfplotstableread[
            col sep=comma,
            header=true,
        ]{
            distribution,cascading,classic
            +10,0,114
            9--10,0,26
            8--9,0,48
            7--8,0,38
            6--7,0,28
            5--6,0,38
            4--5,0,17
            3--4,0,4
            2--3,0,0
            1--2,186,0
        }\loadedtable
        \pgfplotstablecreatecol[
            expr accum={
                round(\pgfmathaccuma) + \thisrow{cascading} + \thisrow{classic}
            }{0}
        ]{sum}{\loadedtable}
        \tikzset{
            fpu=true,
        }
            \pgfplotstablegetrowsof{\loadedtable}
                \pgfmathsetmacro{\LastRow}{\pgfplotsretval-1}
            \pgfplotstablegetelem{9}{sum}\of{\loadedtable}
                \pgfmathsetmacro{\Sum}{\pgfplotsretval}
        \tikzset{
            fpu=false,
        }
        \begin{axis}[
            pyramid axis style,
            %
            axis y line*=left,
            ytick=\empty,
            name=popaxis,
        ]
            \addplot [cyan,fill=cyan!70] table [
                y expr =\coordindex,x expr={\thisrow{classic}/313*100},
            ] \loadedtable;

            \node [anchor=south] at (rel axis cs:0.65,0.6)
                {\textcolor{cyan}{PI-explanation}};
        \end{axis}

        \begin{axis}[
            pyramid axis style,
            %
            at={(popaxis.west)},
            anchor=east,
            xshift=-12.5mm,
            %
            x dir=reverse,
            every node near coord/.append style={
                anchor=east,
            },
            axis y line*=right,
        ]
            \addplot [teal,fill=teal!65] table [
                y expr =\coordindex, x expr={\thisrow{cascading}/186*100},
            ] \loadedtable;

            \node [anchor=south] at (rel axis cs:0.6,0.6)
                {\textcolor{teal}{Cascading Decision Trees}};
        \end{axis}

        \begin{axis}[
            blank pyramid axis style,
            %
            at={(popaxis.west)},
            anchor=east,
            xshift=-12.5mm,
            %
            x dir=reverse,
            axis y line*=right,
            xtick=\empty,
            ytick=data,
            yticklabels from table={\loadedtable}{distribution},
            y tick label style={
                align=center,
                inner sep=0pt,
                text width=12.5mm,
            },
            major tick length=0pt,
            axis line style={
                -,
                draw=none,
            },
        ]
            \addplot [draw=none,fill=none] table [
                y expr =\coordindex, x expr={0},
            ] \loadedtable;

        \end{axis}
    \end{tikzpicture}
\caption{A distribution of explanation lengths for all the examples in the German Credit dataset.}
\label{fig:path_distrib_german}    
\end{figure}
\begin{figure}[h!]
 \centering
      \resizebox{.4\linewidth}{!}{\begin{tikzpicture}
\begin{axis}[
    title={},
    xlabel={Missing data ratio (\%)},
    ylabel={Prediction failure rate (\%)},
    xmin=0, xmax=100,
    ymin=0, ymax=100,
    xtick={0,10,20,30,40,50,60,70,80,90,100},
    ytick={0,10,20,30,40,50,60,70,80,90,100},
    legend pos=south east,
    ymajorgrids=true,
    grid style=dashed,
]

\addplot+[only marks,
    error bars/.cd,
    y dir=both,
    y explicit,
    error bar style={line width=1pt},
    error mark options={
      rotate=90,
      red,
      mark size=4pt,
      line width=0.5pt
    }
    ]
    coordinates {
    (0,0) +- (0,0)
    (4.2,29) +- (0, 2.9)
    (12.5,62.9) +- (0, 1.9)
    (20.8,82.4) +- (0, 4.5)
    (29.2,92.1) +- (0, 0.6)
    (37.5,96.8) +- (0, 1.6)
    (50,99.4) +- (0, 0.6)
    (100,100) +- (0,0)
    };

\addplot+[only marks,
    error bars/.cd,
    y dir=both,
    y explicit,
    error bar style={line width=1pt},
    error mark options={
      rotate=90,
      red,
      mark size=4pt,
      line width=0.5pt
    }
    ]
    coordinates {
    (0,0) +- (0,0)
    (4.2,9.1) +- (0,0.5)
    (12.5,22.6) +- (0,3.8)
    (20.8,37.6) +- (0,2.7)
    (29.2,44.9) +- (0,3.5)
    (37.5,61.9) +- (0,3.3)
    (50,81.7) +- (0,3.3)
    (100,100) +- (0,0)
    };

\addplot[
    color=blue,
    mark=triangle,
    ]
    coordinates {
    (0,0)(4.2,31.9)(12.5,62.9)(20.8,82.4)(29.2,92.1)(37.5,96.8)(50,99.4)(100,100)
    
    };

\addplot[
    color=red,
    mark=square,
    ]
    coordinates {
    (0,0)(4.2,9.1)(12.5,22.6)(20.8,37.6)(29.2,44.9)(37.5,61.9)(50,81.7)(100,100)
    };
    \legend{,,PI-explanation,Cascading Decision Trees}
    
\end{axis}
\end{tikzpicture}}
      \caption{Robustness against missing data. Cascading Decision Tree vs. PI-explanation in the German Credit dataset.}
      \label{fig:linegraph_german}
\end{figure}

\newpage
\subsection{Climate}
\label{subsec:climate}
The goal is to predict climate model simulation outcomes (fail or succeed) given scaled values of climate model input parameters. 

The distribution of explanation lengths for all the examples in the Climate dataset is presented in Figure~\ref{fig:path_distrib_climate}. 
The robustness against missing data is demonstrated in Figure~\ref{fig:linegraph_climate}.


    \pgfplotsset{
        compat=1.9,
        %
        blank pyramid axis style/.style={
            width=0.3*\textwidth,
            height=0.3*\textheight,
            scale only axis,
            xmin=0,
            xmax=100,
            ymin=-0.5,
            ymax=9,
            y dir=reverse,
            enlarge y limits={value=0.075,upper},
            xbar,
            axis x line=left,
            xtick align=outside,
            bar width=1,
            allow reversal of rel axis cs=false,
        },
        pyramid axis style/.style={
            blank pyramid axis style,
            %
            xticklabel={%
                \pgfmathprintnumber\tick\%%
            },
            ytick=\empty,
            axis line style={-},
            %
            nodes near coords={%
                \pgfmathprintnumber\pgfplotspointmeta\%%
            },
            every node near coord/.append style={
                font=\small,
                color=black,
                /pgf/number format/fixed,
            },
        },
    }
\begin{figure}[h!]
\centering
    \begin{tikzpicture}[scale=0.63]
        \pgfplotstableread[
            col sep=comma,
            header=true,
        ]{
            distribution,cascading,classic
            
            +10,0,0
            9--10,0,0
            8--9,0,0
            7--8,0,2
            6--7,0,5
            5--6,0,18
            4--5,0,32
            3--4,0,46
            2--3,0,169
            1--2,502,221
        }\loadedtable
        \pgfplotstablecreatecol[
            expr accum={
                round(\pgfmathaccuma) + \thisrow{cascading} + \thisrow{classic}
            }{0}
        ]{sum}{\loadedtable}
        \tikzset{
            fpu=true,
        }
            \pgfplotstablegetrowsof{\loadedtable}
                \pgfmathsetmacro{\LastRow}{\pgfplotsretval-1}
            \pgfplotstablegetelem{9}{sum}\of{\loadedtable}
                \pgfmathsetmacro{\Sum}{\pgfplotsretval}
        \tikzset{
            fpu=false,
        }
        \begin{axis}[
            pyramid axis style,
            %
            axis y line*=left,
            ytick=\empty,
            name=popaxis,
        ]
            \addplot [cyan,fill=cyan!70] table [
                y expr =\coordindex,x expr={\thisrow{classic}/493*100},
            ] \loadedtable;

            \node [anchor=south] at (rel axis cs:0.65,0.6)
                {\textcolor{cyan}{PI-explanation}};
        \end{axis}

        \begin{axis}[
            pyramid axis style,
            %
            at={(popaxis.west)},
            anchor=east,
            xshift=-12.5mm,
            %
            x dir=reverse,
            every node near coord/.append style={
                anchor=east,
            },
            axis y line*=right,
        ]
            \addplot [teal,fill=teal!65] table [
                y expr =\coordindex, x expr={\thisrow{cascading}/502*100},
            ] \loadedtable;

            \node [anchor=south] at (rel axis cs:0.6,0.6)
                {\textcolor{teal}{Cascading Decision Trees}};
        \end{axis}

        \begin{axis}[
            blank pyramid axis style,
            %
            at={(popaxis.west)},
            anchor=east,
            xshift=-12.5mm,
            %
            x dir=reverse,
            axis y line*=right,
            xtick=\empty,
            ytick=data,
            yticklabels from table={\loadedtable}{distribution},
            y tick label style={
                align=center,
                inner sep=0pt,
                text width=12.5mm,
            },
            major tick length=0pt,
            axis line style={
                -,
                draw=none,
            },
        ]
            \addplot [draw=none,fill=none] table [
                y expr =\coordindex, x expr={0},
            ] \loadedtable;

        \end{axis}
    \end{tikzpicture}
\caption{A distribution of explanation lengths for all the examples in the Climate dataset.}
\label{fig:path_distrib_climate}    
\end{figure}

\begin{figure}[h!]
 \centering
      \resizebox{.4\linewidth}{!}{\begin{tikzpicture}
\begin{axis}[
    title={},
    xlabel={Missing data ratio (\%)},
    ylabel={Prediction failure rate (\%)},
    xmin=0, xmax=100,
    ymin=0, ymax=100,
    xtick={0,10,20,30,40,50,60,70,80,90,100},
    ytick={0,10,20,30,40,50,60,70,80,90,100},
    legend pos=south east,
    ymajorgrids=true,
    grid style=dashed,
]

\addplot+[only marks,
    error bars/.cd,
    y dir=both,
    y explicit,
    error bar style={line width=1pt},
    error mark options={
      rotate=90,
      red,
      mark size=4pt,
      line width=0.5pt
    }
    ]
    coordinates {
    (0,0) +- (0,0)
    (5.5,14.2) +- (0,2.2)
    (11,28.0) +- (0,2.2)
    (16.5,41.8) +- (0,0.2)
    (27.5,63.5) +- (0,0.6)
    (44.4,79.5) +- (0,1.4)
    (66.7,94.3) +- (0,1.0)
    (100,100) +- (0,0)
    };

\addplot+[only marks,
    error bars/.cd,
    y dir=both,
    y explicit,
    error bar style={line width=1pt},
    error mark options={
      rotate=90,
      red,
      mark size=4pt,
      line width=0.5pt
    }
    ]
    coordinates {
    (0,0) +- (0,0)
    (5.5,10.2) +- (0,1.6)
    (11,21.9) +- (0,2.6)
    (16.5,31.4) +- (0,1)
    (27.5,47.8) +- (0,1.6)
    (44.4,71.9) +- (0,2.2)
    (66.7,89.8) +- (0,0.4)
    (100,100) +- (0,0)
    };

\addplot[
    color=blue,
    mark=triangle,
    ]
    coordinates {
    (0,0)(5.5,14.2)(11,28.0)(16.5,41.8)(27.5,63.5)(44.4,79.5)(66.7,94.3)(100,100)

    };

\addplot[
    color=red,
    mark=square,
    ]
    coordinates {
    (0,0)(5.5,10.2)(11,21.9)(16.5,31.4)(27.5,47.8)(44.4,71.9)(66.7,89.8)(100,100)
    };
    \legend{,,PI-explanation,Cascading Decision Trees}
    
\end{axis}
\end{tikzpicture}}
      \caption{Robustness against missing data. Cascading Decision Tree vs. PI-explanation in the Climate dataset.}
      \label{fig:linegraph_climate}
\end{figure}

\newpage
\subsection{Happiness}
\label{subsec:happy}
Pridicting the happiness of people using the Receiver Operating Characteristic (ROC).

The distribution of explanation lengths for all the examples in the Happiness dataset is presented in Figure~\ref{fig:path_distrib_happy}. 
The robustness against missing data is demonstrated in Figure~\ref{fig:linegraph_happy}.


    \pgfplotsset{
        compat=1.9,
        %
        blank pyramid axis style/.style={
            width=0.3*\textwidth,
            height=0.3*\textheight,
            scale only axis,
            xmin=0,
            xmax=100,
            ymin=-0.5,
            ymax=9,
            y dir=reverse,
            enlarge y limits={value=0.075,upper},
            xbar,
            axis x line=left,
            xtick align=outside,
            bar width=1,
            allow reversal of rel axis cs=false,
        },
        pyramid axis style/.style={
            blank pyramid axis style,
            %
            xticklabel={%
                \pgfmathprintnumber\tick\%%
            },
            ytick=\empty,
            axis line style={-},
            %
            nodes near coords={%
                \pgfmathprintnumber\pgfplotspointmeta\%%
            },
            every node near coord/.append style={
                font=\small,
                color=black,
                /pgf/number format/fixed,
            },
        },
    }
\begin{figure}[h!]
\centering
    \begin{tikzpicture}[scale=0.63]
        \pgfplotstableread[
            col sep=comma,
            header=true,
        ]{
            distribution,cascading,classic
            +10,0,14
            9--10,0,4
            8--9,0,5
            7--8,0,15
            6--7,0,1
            5--6,0,6
            4--5,0,7
            3--4,0,12
            2--3,0,13
            1--2,46,0

        }\loadedtable
        \pgfplotstablecreatecol[
            expr accum={
                round(\pgfmathaccuma) + \thisrow{cascading} + \thisrow{classic}
            }{0}
        ]{sum}{\loadedtable}
        \tikzset{
            fpu=true,
        }
            \pgfplotstablegetrowsof{\loadedtable}
                \pgfmathsetmacro{\LastRow}{\pgfplotsretval-1}
            \pgfplotstablegetelem{9}{sum}\of{\loadedtable}
                \pgfmathsetmacro{\Sum}{\pgfplotsretval}
        \tikzset{
            fpu=false,
        }
        \begin{axis}[
            pyramid axis style,
            %
            axis y line*=left,
            ytick=\empty,
            name=popaxis,
        ]
            \addplot [cyan,fill=cyan!70] table [
                y expr =\coordindex,x expr={\thisrow{classic}/77*100},
            ] \loadedtable;

            \node [anchor=south] at (rel axis cs:0.65,0.6)
                {\textcolor{cyan}{PI-explanation}};
        \end{axis}

        \begin{axis}[
            pyramid axis style,
            %
            at={(popaxis.west)},
            anchor=east,
            xshift=-12.5mm,
            %
            x dir=reverse,
            every node near coord/.append style={
                anchor=east,
            },
            axis y line*=right,
        ]
            \addplot [teal,fill=teal!65] table [
                y expr =\coordindex, x expr={\thisrow{cascading}/46*100},
            ] \loadedtable;

            \node [anchor=south] at (rel axis cs:0.6,0.6)
                {\textcolor{teal}{Cascading Decision Trees}};
        \end{axis}

        \begin{axis}[
            blank pyramid axis style,
            %
            at={(popaxis.west)},
            anchor=east,
            xshift=-12.5mm,
            %
            x dir=reverse,
            axis y line*=right,
            xtick=\empty,
            ytick=data,
            yticklabels from table={\loadedtable}{distribution},
            y tick label style={
                align=center,
                inner sep=0pt,
                text width=12.5mm,
            },
            major tick length=0pt,
            axis line style={
                -,
                draw=none,
            },
        ]
            \addplot [draw=none,fill=none] table [
                y expr =\coordindex, x expr={0},
            ] \loadedtable;

        \end{axis}
    \end{tikzpicture}
\caption{A distribution of explanation lengths for all the examples in the Happiness dataset.}
\label{fig:path_distrib_happy}    
\end{figure}

\begin{figure}[h!]
 \centering
      \resizebox{.4\linewidth}{!}{\begin{tikzpicture}
\begin{axis}[
    title={},
    xlabel={Missing data ratio (\%)},
    ylabel={Prediction failure rate (\%)},
    xmin=0, xmax=100,
    ymin=0, ymax=100,
    xtick={0,10,20,30,40,50,60,70,80,90,100},
    ytick={0,10,20,30,40,50,60,70,80,90,100},
    legend pos=south east,
    ymajorgrids=true,
    grid style=dashed,
]

\addplot+[only marks,
    error bars/.cd,
    y dir=both,
    y explicit,
    error bar style={line width=1pt},
    error mark options={
      rotate=90,
      red,
      mark size=4pt,
      line width=0.5pt
    }
    ]
    coordinates {
    (0,0) +- (0,0)
    (16.7,68.8) +- (0,2.6)
    (33.3,90) +- (0,2.6)
    (50,97.4) +- (0,2.6)
    (66.7,100) +- (0,0.0)
    (100,100) +- (0,0)
    };

\addplot+[only marks,
    error bars/.cd,
    y dir=both,
    y explicit,
    error bar style={line width=1pt},
    error mark options={
      rotate=90,
      red,
      mark size=4pt,
      line width=0.5pt
    }
    ]
    coordinates {
    (0,0) +- (0,0)
    (16.7,34.8) +- (0,4.3)
    (33.3,60.9) +- (0,4.3)
    (50,76.1) +- (0,6.5)
    (66.7,91.3) +- (0,4.3)
    (100,100) +- (0,0)
    };

\addplot[
    color=blue,
    mark=triangle,
    ]
    coordinates {
    (0,0)(16.7,68.8)(33.3,90)(50,97.4)(66.7,100)(100,100)
    
    };

\addplot[
    color=red,
    mark=square,
    ]
    coordinates {
    (0,0)(16.7,34.8)(33.3,60.9)(50,76.1)(66.7,91.3)(100,100)
    };
    \legend{,,PI-explanation,Cascading Decision Trees}
    
\end{axis}
\end{tikzpicture}}
      \caption{Robustness against missing data. Cascading Decision Tree vs. PI-explanation in the Happiness dataset.}
      \label{fig:linegraph_happy}
\end{figure}

\newpage
\subsection{Ionosphere}
\label{subsec:iono}
Detection of free electrons in the ionosphere. ``Positive'' samples show the detecting signals fail to detect the free electrons; the signals just pass through the ionosphere. ``Negative'' samples show some evidence of the stucture in the ionosphere. 

The distribution of explanation lengths for all the examples in the Ionosphere dataset is presented in Figure~\ref{fig:path_distrib_iono}. 
The robustness against missing data is demonstrated in Figure~\ref{fig:linegraph_iono}.


    \pgfplotsset{
        compat=1.9,
        %
        blank pyramid axis style/.style={
            width=0.3*\textwidth,
            height=0.3*\textheight,
            scale only axis,
            xmin=0,
            xmax=100,
            ymin=-0.5,
            ymax=9,
            y dir=reverse,
            enlarge y limits={value=0.075,upper},
            xbar,
            axis x line=left,
            xtick align=outside,
            bar width=1,
            allow reversal of rel axis cs=false,
        },
        pyramid axis style/.style={
            blank pyramid axis style,
            %
            xticklabel={%
                \pgfmathprintnumber\tick\%%
            },
            ytick=\empty,
            axis line style={-},
            %
            nodes near coords={%
                \pgfmathprintnumber\pgfplotspointmeta\%%
            },
            every node near coord/.append style={
                font=\small,
                color=black,
                /pgf/number format/fixed,
            },
        },
    }
\begin{figure}[h!]
\centering
    \begin{tikzpicture}[scale=0.63]
        \pgfplotstableread[
            col sep=comma,
            header=true,
        ]{
            distribution,cascading,classic
            +10,0,0
            9--10,0,0
            8--9,0,0
            7--8,0,0
            6--7,0,6
            5--6,0,8
            4--5,0,11
            3--4,0,17
            2--3,0,12
            1--2,117,69

        }\loadedtable
        \pgfplotstablecreatecol[
            expr accum={
                round(\pgfmathaccuma) + \thisrow{cascading} + \thisrow{classic}
            }{0}
        ]{sum}{\loadedtable}
        \tikzset{
            fpu=true,
        }
            \pgfplotstablegetrowsof{\loadedtable}
                \pgfmathsetmacro{\LastRow}{\pgfplotsretval-1}
            \pgfplotstablegetelem{9}{sum}\of{\loadedtable}
                \pgfmathsetmacro{\Sum}{\pgfplotsretval}
        \tikzset{
            fpu=false,
        }
        \begin{axis}[
            pyramid axis style,
            %
            axis y line*=left,
            ytick=\empty,
            name=popaxis,
        ]
            \addplot [cyan,fill=cyan!70] table [
                y expr =\coordindex,x expr={\thisrow{classic}/124*100},
            ] \loadedtable;

            \node [anchor=south] at (rel axis cs:0.65,0.6)
                {\textcolor{cyan}{PI-explanation}};
        \end{axis}

        \begin{axis}[
            pyramid axis style,
            %
            at={(popaxis.west)},
            anchor=east,
            xshift=-12.5mm,
            %
            x dir=reverse,
            every node near coord/.append style={
                anchor=east,
            },
            axis y line*=right,
        ]
            \addplot [teal,fill=teal!65] table [
                y expr =\coordindex, x expr={\thisrow{cascading}/117*100},
            ] \loadedtable;

            \node [anchor=south] at (rel axis cs:0.6,0.6)
                {\textcolor{teal}{Cascading Decision Trees}};
        \end{axis}

        \begin{axis}[
            blank pyramid axis style,
            %
            at={(popaxis.west)},
            anchor=east,
            xshift=-12.5mm,
            %
            x dir=reverse,
            axis y line*=right,
            xtick=\empty,
            ytick=data,
            yticklabels from table={\loadedtable}{distribution},
            y tick label style={
                align=center,
                inner sep=0pt,
                text width=12.5mm,
            },
            major tick length=0pt,
            axis line style={
                -,
                draw=none,
            },
        ]
            \addplot [draw=none,fill=none] table [
                y expr =\coordindex, x expr={0},
            ] \loadedtable;

        \end{axis}
    \end{tikzpicture}
\caption{A distribution of explanation lengths for all the examples in the Ionosphere dataset.}
\label{fig:path_distrib_iono}    
\end{figure}
\begin{figure}[h!]
 \centering
      \resizebox{.4\linewidth}{!}{\begin{tikzpicture}
\begin{axis}[
    title={},
    xlabel={Missing data ratio (\%)},
    ylabel={Prediction failure rate (\%)},
    xmin=0, xmax=100,
    ymin=0, ymax=100,
    xtick={0,10,20,30,40,50,60,70,80,90,100},
    ytick={0,10,20,30,40,50,60,70,80,90,100},
    legend pos=south east,
    ymajorgrids=true,
    grid style=dashed,
]

\addplot+[only marks,
    error bars/.cd,
    y dir=both,
    y explicit,
    error bar style={line width=1pt},
    error mark options={
      rotate=90,
      red,
      mark size=4pt,
      line width=0.5pt
    }
    ]
    coordinates {
    (0,0) +- (0,0)
    (2.9,9.8) +- (0, 0.9)
    (8.8,20.3) +- (0, 2.4)
    (17.6,30.5) +- (0, 6.1)
    (29.4,54.5) +- (0,1.5)
    (47.1,70.7) +- (0,0.8)
    (73.5,85.3) +- (0,4.1)
    (88.2,95.9) +- (0,0.8)
    (100,100) +- (0,0)
    };

\addplot+[only marks,
    error bars/.cd,
    y dir=both,
    y explicit,
    error bar style={line width=1pt},
    error mark options={
      rotate=90,
      red,
      mark size=4pt,
      line width=0.5pt
    }
    ]
    coordinates {
    (0,0) +- (0,0)
    (2.9,1.7) +- (0, 0.9)
    (8.8,11.5) +- (0,3)
    (17.6,24.8) +- (0, 0.9)
    (29.4,32.5) +- (0,2.5)
    (47.1,52.1) +- (0,2.5)
    (73.5,76.1) +- (0,2.5)
    (88.2,90.6) +- (0,2.6)
    (100,100) +- (0,0)
    };

\addplot[
    color=blue,
    mark=triangle,
    ]
    coordinates {
    (0,0)(2.9,9.8)(8.8,20.3)(17.6,30.5)(29.4,54.5)(47.1,70.7)(73.5,85.3)(88.2,95.9)(100,100)
    
    };

\addplot[
    color=red,
    mark=square,
    ]
    coordinates {
    (0,0)(2.9,0.9)(8.8,14.5)(17.6,24.8)(29.4,32.5)(47.1,52.1)(73.5,76.1)(88.2,90.6)(100,100)
    };
    \legend{,,PI-explanation,Cascading Decision Trees}
    
\end{axis}
\end{tikzpicture}}
      \caption{Robustness against missing data. Cascading Decision Tree vs. PI-explanation in the Ionosphere dataset.}
      \label{fig:linegraph_iono}
\end{figure}

\newpage
\subsection{Sonar}
\label{subsec:sonar}
Discrimination of objects. This dataset includes bouncing sonar signals off a mine (metal cylinder) at various angles and under various conditions. ``Positive'' samples indicate the object is indeed a rock not a mine, while ``Negative'' samples indicate the object is a mine.

The distribution of explanation lengths for all the examples in the Sonar dataset is presented in Figure~\ref{fig:path_distrib_sonar}. 
The robustness against missing data is demonstrated in Figure~\ref{fig:linegraph_sonar}.


    \pgfplotsset{
        compat=1.9,
        %
        blank pyramid axis style/.style={
            width=0.3*\textwidth,
            height=0.3*\textheight,
            scale only axis,
            xmin=0,
            xmax=100,
            ymin=-0.5,
            ymax=9,
            y dir=reverse,
            enlarge y limits={value=0.075,upper},
            xbar,
            axis x line=left,
            xtick align=outside,
            bar width=1,
            allow reversal of rel axis cs=false,
        },
        pyramid axis style/.style={
            blank pyramid axis style,
            %
            xticklabel={%
                \pgfmathprintnumber\tick\%%
            },
            ytick=\empty,
            axis line style={-},
            %
            nodes near coords={%
                \pgfmathprintnumber\pgfplotspointmeta\%%
            },
            every node near coord/.append style={
                font=\small,
                color=black,
                /pgf/number format/fixed,
            },
        },
    }
\begin{figure}[h!]
\centering
    \begin{tikzpicture}[scale=0.63]
        \pgfplotstableread[
            col sep=comma,
            header=true,
        ]{
            distribution,cascading,classic
            +10,0,0
            9--10,0,0
            8--9,0,0
            7--8,0,0
            6--7,0,1
            5--6,0,10
            4--5,0,30
            3--4,0,21
            2--3,0,33
            1--2,82,22
        }\loadedtable
        \pgfplotstablecreatecol[
            expr accum={
                round(\pgfmathaccuma) + \thisrow{cascading} + \thisrow{classic}
            }{0}
        ]{sum}{\loadedtable}
        \tikzset{
            fpu=true,
        }
            \pgfplotstablegetrowsof{\loadedtable}
                \pgfmathsetmacro{\LastRow}{\pgfplotsretval-1}
            \pgfplotstablegetelem{9}{sum}\of{\loadedtable}
                \pgfmathsetmacro{\Sum}{\pgfplotsretval}
        \tikzset{
            fpu=false,
        }
        \begin{axis}[
            pyramid axis style,
            %
            axis y line*=left,
            ytick=\empty,
            name=popaxis,
        ]
            \addplot [cyan,fill=cyan!70] table [
                y expr =\coordindex,x expr={\thisrow{classic}/117*100},
            ] \loadedtable;

            \node [anchor=south] at (rel axis cs:0.65,0.6)
                {\textcolor{cyan}{PI-explanation}};
        \end{axis}

        \begin{axis}[
            pyramid axis style,
            %
            at={(popaxis.west)},
            anchor=east,
            xshift=-12.5mm,
            %
            x dir=reverse,
            every node near coord/.append style={
                anchor=east,
            },
            axis y line*=right,
        ]
            \addplot [teal,fill=teal!65] table [
                y expr =\coordindex, x expr={\thisrow{cascading}/82*100},
            ] \loadedtable;

            \node [anchor=south] at (rel axis cs:0.6,0.6)
                {\textcolor{teal}{Cascading Decision Trees}};
        \end{axis}

        \begin{axis}[
            blank pyramid axis style,
            %
            at={(popaxis.west)},
            anchor=east,
            xshift=-12.5mm,
            %
            x dir=reverse,
            axis y line*=right,
            xtick=\empty,
            ytick=data,
            yticklabels from table={\loadedtable}{distribution},
            y tick label style={
                align=center,
                inner sep=0pt,
                text width=12.5mm,
            },
            major tick length=0pt,
            axis line style={
                -,
                draw=none,
            },
        ]
            \addplot [draw=none,fill=none] table [
                y expr =\coordindex, x expr={0},
            ] \loadedtable;

        \end{axis}
    \end{tikzpicture}
\caption{A distribution of explanation lengths for all the examples in the Sonar dataset.}
\label{fig:path_distrib_sonar}    
\end{figure}
\begin{figure}[h!]
 \centering
      \resizebox{.4\linewidth}{!}{\begin{tikzpicture}
\begin{axis}[
    title={},
    xlabel={Missing data ratio (\%)},
    ylabel={Prediction failure rate (\%)},
    xmin=0, xmax=100,
    ymin=0, ymax=100,
    xtick={0,10,20,30,40,50,60,70,80,90,100},
    ytick={0,10,20,30,40,50,60,70,80,90,100},
    legend pos=south east,
    ymajorgrids=true,
    grid style=dashed,
]

\addplot+[only marks,
    error bars/.cd,
    y dir=both,
    y explicit,
    error bar style={line width=1pt},
    error mark options={
      rotate=90,
      red,
      mark size=4pt,
      line width=0.5pt
    }
    ]
    coordinates {
    (0,0) +- (0,0)
    (1.7,3.8) +- (0,1.2)
    (6.7,19.7) +- (0,1.8)
    (13.3,39.7) +- (0,4.1)
    (23.3,61.1) +- (0,3.0)
    (36.7,84.6) +- (0,4.3)
    (58.3,95.3) +- (0,2.1)
    (75,98.3) +- (0,0)
    (100,100) +- (0,0)
    };

\addplot+[only marks,
    error bars/.cd,
    y dir=both,
    y explicit,
    error bar style={line width=1pt},
    error mark options={
      rotate=90,
      red,
      mark size=4pt,
      line width=0.5pt
    }
    ]
    coordinates {
    (0,0) +- (0,0)
    (1.7,1.8) +- (0,0.6)
    (6.7,11.0) +- (0,2.6)
    (13.3,24.4) +- (0,4.9)
    (23.3,43.3) +- (0,3.0)
    (36.7,65.5) +- (0,7.8)
    (58.3,78.0) +- (0,3.6)
    (75,92.0) +- (0,2.0)
    (100,100) +- (0,0)
    };

\addplot[
    color=blue,
    mark=triangle,
    ]
    coordinates {
    (0,0)(1.7,3.8)(6.7,19.7)(13.3,39.7)(23.3,61.1)(36.7,84.6)(58.3,97.4)(75,98.3)(100,100)
    
    };

\addplot[
    color=red,
    mark=square,
    ]
    coordinates {
    (0,0)(1.7,1.8)(6.7,11.0)(13.3,24.4)(23.3,43.3)(36.7,65.5)(58.3,81.7)(75,93.9)(100,100)
    };
    \legend{,,PI-explanation,Cascading Decision Trees}
    
\end{axis}
\end{tikzpicture}}
      \caption{Robustness against missing data. Cascading Decision Tree vs. PI-explanation in the Sonar dataset.}
      \label{fig:linegraph_sonar}
\end{figure}

\end{document}